\documentclass[times,twocolumn,final]{elsarticle}

\usepackage{medima}
\usepackage{framed,multirow}

\usepackage{amssymb}
\usepackage{latexsym}

\usepackage{url}
\usepackage{xcolor}

\usepackage{hyperref}

\usepackage{graphicx}
\usepackage{array}
\usepackage{multirow}
\usepackage{xcolor}
\usepackage{threeparttable}
\usepackage{booktabs} 
\usepackage{multicol}
\usepackage{algorithm}
\usepackage{bm}
\usepackage{verbatim}
\usepackage{graphicx}
\usepackage{cite}
\usepackage{xcolor}
\usepackage{bbding}
\usepackage{tabularx}
\usepackage{makecell}
\usepackage{amsmath}

\definecolor{newcolor}{rgb}{.8,.349,.1}

\journal{Medical Image Analysis}

\begin{document}

\verso{Peng Ling}

\begin{frontmatter}

\title{FrGNet: A fourier-guided weakly-supervised framework for nuclear instance segmentation}%

\author[1]{Peng \snm{Ling}\fnref{fn1}}
\author[2]{Wenxiao \snm{Xiong}}
\fntext[fn2]{Wenxiao Xiong: The School of Computer Science and Engineering, Sun Yat-sen University, Guangzhou, China(xiongwx6@mail2.sysu.edu.cn).}
\cortext[cor1]{Corresponding author at: 
  Personal.
  \textit{E-mail address}: lingp1999@gmail.com.
  }


\begin{abstract}
Nuclear instance segmentation has played a critical role in pathology image analysis. 
The main challenges arise from the difficulty in accurately segmenting instances and the high cost of precise mask-level annotations for fully-supervised training.
In this work, we propose a fourier guidance framework for solving the weakly-supervised nuclear instance segmentation problem.
In this framework, we construct a fourier guidance module to fuse the priori information into the training process of the model, which facilitates the model to capture the relevant features of the nuclear.
Meanwhile, in order to further improve the model's ability to represent the features of nuclear, we propose the guide-based instance level contrastive module.
This module makes full use of the framework's own properties and guide information to effectively enhance the representation features of nuclear.
We show on two public datasets that our model can outperform current SOTA methods under fully-supervised design, and in weakly-supervised experiments, with only a small amount of labeling our model still maintains close to the performance under full supervision.
In addition, we also perform generalization experiments on a private dataset, and without any labeling, our model is able to segment nuclear images that have not been seen during training quite effectively.
 As open science, all codes and pre-trained
models are available at https://github.com/LQY404/FrGNet.
\end{abstract}

\begin{keyword}
\KWD \\ Nuclear instance segmentation \\ Weakly-supervised learning \\ Deep learning \\ Medical image segmentation
\end{keyword}

\end{frontmatter}


\section{Introduction}
Pathological slide analysis is widely regarded as the gold standard for cancer diagnosis, treatment, and prevention.
Nuclear instance segmentation is a critical step in this process, because the nuclear features such
as average size, density and nucleus-to-cytoplasm ratio are related to the clinical diagnosis and management of cancer.
In general, some of the major nuclear segmentation methods are based on fully-supervised designs~\citep{latorre2013segmentation, abbas2014occluded, sheeba2014splitting, zhou2019cia, chen2017dcan, graham2019hover, he2017mask, liu2021panoptic, naylor2018segmentation, he2021hybrid, qu2019improving, raza2019micro, ronneberger2015u, chen2023cpp, liu2021mdc, zhou2022semantic, upschulte2022contour, he2023toposeg}, which can be divided into three main categories: 1) methods based on traditional image manipulation~\citep{latorre2013segmentation, abbas2014occluded, sheeba2014splitting}, which rely on cumbersome post-processing processes; 2) detection-based methods, which either use a strategy of detecting and then segmenting the algorithms~\citep{he2017mask}, or model the nuclear using contour and regress them directly to to obtain results~\citep{upschulte2022contour}; 3) segmentation-based methods, which directly generate multiple segmentation masks and then use post-processing operations to obtain nuclear segmentation instances~\citep{zhou2019cia, chen2017dcan, graham2019hover, he2017mask, liu2021panoptic, naylor2018segmentation, qu2019improving, raza2019micro, ronneberger2015u, chen2023cpp, liu2021mdc, zhou2022semantic, he2023transnuseg, nam2023pronet, he2023toposeg}.
However, the fully supervised learning of deep neural networks in these methods requires a large amount of training data, which are pixel-wise annotated. 
It is difficult to collect such datasets because assigning a nuclear/background class label to every pixel in the image is very time-consuming and requires expert domain knowledge.

\begin{figure*}[tb]
\centering
  \includegraphics[width=0.99\linewidth]{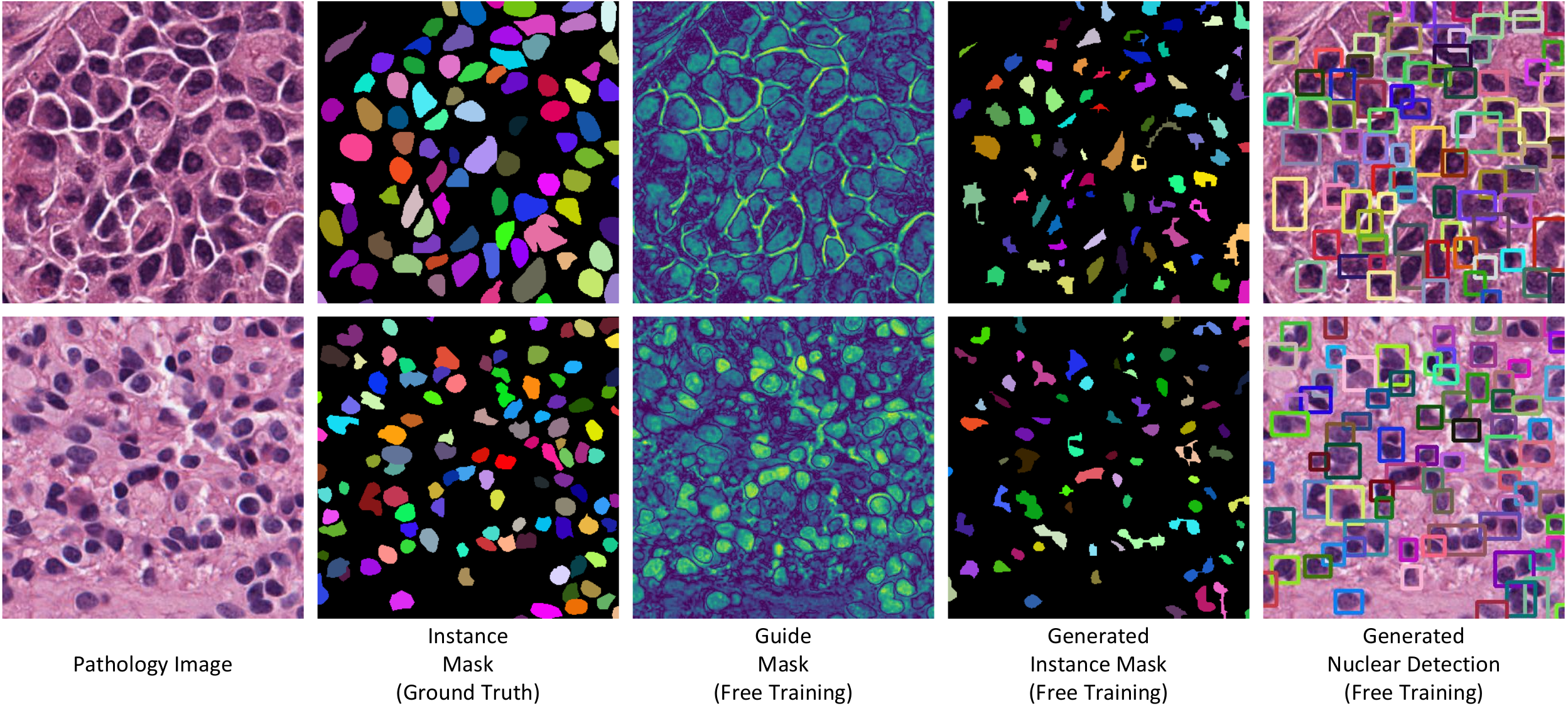}
  \caption{
  Motivations. We can get corse nuclear instance segmentation results without any training.
  }
\label{fig:teaser}
\end{figure*}

In this work, we find that as shown in Fig.~\ref{fig:teaser}, we can get coarse nuclear instance segmentation without any training.
Specifically, for a nuclear pathology image, the approximate location of the nuclear can be obtained by fourier transform (guide mask).
Obviously, such a guide mask can provide very powerful a priori information for nuclear segmentation.
In order to fully utilize this a priori information
We introduce a framework for segmenting nuclear instances using fourier guidance. 
This framework integrates intrinsic feature information from nuclear images and incorporates a priori information about nuclear locations through a specially designed Fourier Guidance (FG) module, which directs the training process.
To further enhance model performance, we propose a Guide-based Instance Level Contrastive (GILC) module. This module offers instance-level feature guidance and strengthens the feature representation of nuclear instances.

The contributions can be summarized as follows:
\begin{itemize}
    \item We propose a new framework FrGNet, for nuclei instance segmentation in histopathologic images. 
    This framework takes the characteristic of nuclear into consideration, utilizes masks generated by Fourier transformation as guidance to effectively address both full-supervised and weak-supervised nuclei instance segmentation tasks.
    \item We propose the Fourier Guidance (\textbf{FG}) Module. This module make full use of the characteristics of nuclear image, providing the model with a priori information about the location of the nuclear, allowing the model to better localise it.
    \item To further improve the model performance, we propose the guide-based instance level contrastive (\textbf{GILC}) module, which provides instance-level feature guidance for the model and enhances the feature representation of the nuclear instances.
    \item We demonstrate the efficiency of our proposed framework through experiments on two classical nuclei instance segmentation datasets. 
    Compared to recent methods, our approach achieves superior performance, setting a new state-of-the-art (SOTA) benchmark. 
    Additionally, we showcase the capability of our framework in the field of weak-supervised nuclei instance segmentation.
    Furthermore, the powerful generalization of our model is demonstrated by generalization experiments on a private dataset \textbf{without any annotations}.
\end{itemize}

\section{Related Work}

\subsection{Nuclear instance segmentation}
Before the emergence of deep learning, instance segmentation methods predominantly relied on classical machine learning algorithms and image processing techniques. 
These included statistical features, thresholding-based methods~\citep{latorre2013segmentation, abbas2014occluded, sheeba2014splitting}, and morphological features for contour and shape identification. 
Energy-based solutions gained prominence during the 1990s. 
Marker-based approaches such as the Watershed algorithm proved beneficial for nuclear instance segmentation, as demonstrated by \citet{cheng2008segmentation}. 
However, these techniques faced limitations in accurately segmenting nuclear due to their diverse structures, textures, intensities, leading to unreliable outcomes. 
Moreover, the effectiveness of these systems heavily depended on manual parameter tuning, including thresholds and weights, rendering them unsuitable for widespread application due to their inherent unreliability.

With the rapid development of deep learning, \citet{ronneberger2015u} proposed the UNet model, which has since become one of the most fundamental models in medical image segmentation. 
\citet{raza2019micro} introduced Micro-Net, achieving robustness to large internal and external variances in nuclear size by utilizing multi-resolution and weighted loss functions. \citet{qu2019improving} developed a full-resolution convolutional neural network (FullNet), which enhances localization accuracy by eliminating downsampling operations in the network structure. 
\citet{he2021hybrid} presented a hybrid attention nested U-shaped network (Han-Net) to extract effective feature information from multiple layers.

To leverage contour information for distinguishing contact/overlapping nuclear, \citet{chen2017dcan} initially proposed incorporating contour information into a multi-level fully convolutional network (FCN) to create a deep contour-aware network for nuclear instance segmentation. Subsequently,
\citet{zhou2019cia} introduced the contour-aware information aggregation network, which combines spatial and textural features between nuclear and contours. 
Additionally, some models have approached nuclear instance segmentation as an object detection task, such as contour proposal networks (CPN)~\citep{upschulte2022contour}, which use a sparse list of contour representations to define a nuclear instance.

Several works~\citep{chen2023cpp, graham2019hover, liu2021mdc, naylor2018segmentation} have introduced distance maps to separate contact/overlapping nuclear. 
\citet{naylor2018segmentation} addressed the issue of segmenting touching nuclear by formulating the segmentation task as a regression task of intra-nuclear distance maps.
\citet{graham2019hover} proposed Hover-Net, a network for simultaneous nuclear segmentation and classification, which uses the vertical and horizontal distances between a nuclear pixel and its center of mass to separate clusters of nuclear. 
Moreover, \citet{he2021cdnet} introduced a centripetal directional network (CDNet) for nuclear instance segmentation, incorporating directional information into the network. \citet{he2023toposeg} take topological information into consideration to further split overlapping nuclear instance.
These works mainly are proposed for full-supervised nuclear instance segmentation, and don't consider the characteristic of nuclear itself.
In this work, we make full use of the characteristic of nuclear instance (we name it as \textbf{nuclear guidance}), introduce a new framework with guidance, not only can solve full-supervised but also weak-supervised nuclear instance segmentation efficiently.

\subsection{Weakly-supervised segmentation}
Weakly supervised approaches offer the advantage of reducing manual annotation effort compared to fully supervised methods. 
In natural image segmentation, ~\citet{papandreou2015weakly} proposed an Expectation-Maximization (EM) method for training with image-level or bounding-box annotations. ~\citet{pathak2015constrained} added a set of linear constraints on the output space in the loss function to leverage information from image-level labels.

In contrast to image-level annotations, point annotations provide more precise location information for each object. ~\citet{bearman2016s} incorporated an objectness prior in the loss function to guide the training of a CNN, aiding in the separation of objects from the background. Scribble annotations, which require at least one scribble per object, are a more informative type of weak label. ~\citet{lin2016scribblesup} used scribble annotations to train a graphical model that propagates information from the scribbles to the unmarked pixels.

Bounding boxes are the most widely used form of weak annotation, applied in both natural images~\citep{dai2015boxsup, rajchl2016deepcut} and medical images~\citep{yang2018boxnet, zhao2018deep}. 
~\citet{kervadec2019constrained} utilized a small fraction of full labels and imposed a size constraint in their loss function, achieving good performance, though this method is not applicable to scenarios involving multiple objects of the same class. Another method~\citep{qu2020weakly} proposed a two-stage approach that uses only a small fraction of nuclear locations.

In this work, we propose a novel end-to-end framework to solve both fully-supervised and weakly-supervised nuclear instance segmentation tasks.
We start from the characteristics of the nuclear image itself, and use the information of this feature as the a priori information to build the corresponding module to guide the model for training.
With only a small amount of labeled data, our model is able to approach fully-supervised results.

\subsection{Contrastive learning}
Contrastive learning is a highly regarded technique for learning representations from unlabeled features these days~\citep{chen2020simple, chen2020improved, grill2020bootstrap}. 
It aims to enhance representation learning by contrasting similar features (positive pairs) against dissimilar features (negative pairs). A key innovation in contrastive learning lies in the selection of positive and negative pairs. Additionally, the use of a memory bank to store more negative samples has been adopted, as this can lead to improved performance~\citep{chen2020simple}.

In the field of segmentation, numerous works leverage contrastive learning for the pre-training of models~\citep{chaitanya2020contrastive, wang2021dense, xie2021propagate}. 
Recently, ~\citet{wang2021exploring} demonstrated the advantages of applying contrastive learning in a cross-image pixel-wise manner for supervised segmentation. 
The CAC approach~\citep{lai2021semi} shows improvement in semi-supervised segmentation by performing directional contrastive learning pixel-to-pixel, aligning lower-quality features towards their high-quality counterparts.

Unlike these works, we construct the guide-based insatnce level contrastive (GILC) module from the image characteristics of the nuclear, which relies on the automatically generated guide mask to further enhance the feature representation of the nuclear, and is able to be applied to both fully-supervised and weakly-supervised nuclear instance segmentation tasks.

\section{Method}

\subsection{Overview}
\begin{figure*}[tb]
\centering
  \includegraphics[width=0.99\linewidth]{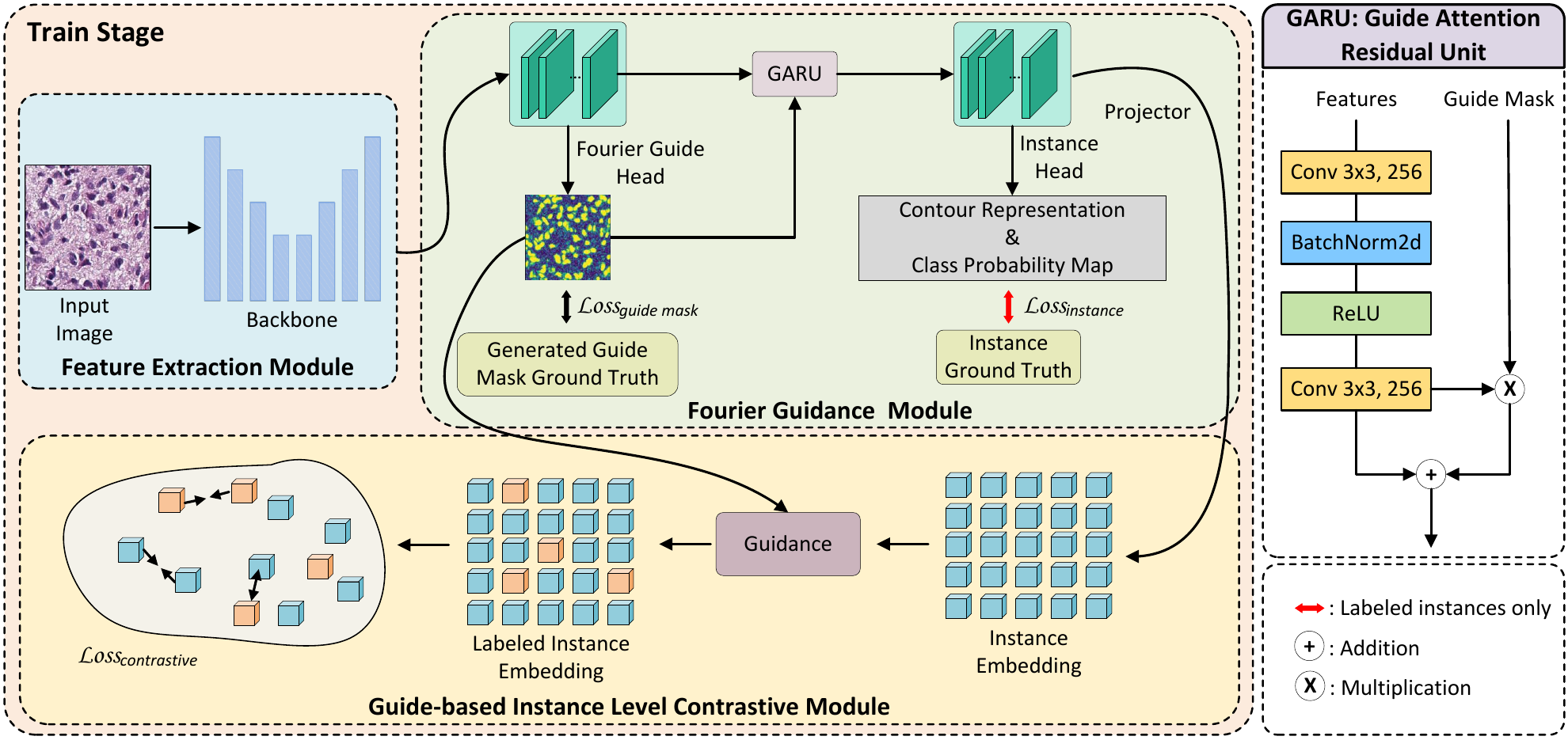}
  \caption{
  The workflow of our proposed FrGNet framework. For an input image, we first extract the image features using the feature extraction module to obtain multi-level image features, then use the features maps to generate guide mask and nuclear instances in fourier guidance module. Furthermore, the proposed instance-level comparison module also make use of the feature maps to enhances the feature representation of the nuclear instances.
  }
\label{fig:framework_train}
\end{figure*}

As illustrated in Fig.~\ref{fig:framework_train}, our framework consist of three parts: feature extraction module, fourier guidance module, and guide-based instance level contrastive module.
For an input nuclear pathology image, we first extract the image features using the feature extraction module to obtain multi-level image features, with the layers being smaller at higher resolutions.
Secondly, for the multilevel image features, we uniformly input them into the fourier guidance module for processing.

In the fourier guidance module, we select the feature map of layer 0 as the input to the fourier guide head to get the predicted guide mask, and use the pre-generated guide mask ground truth for supervision.
Subsequently, for the predicted guide mask, we use guide attention residual unit (GARU) to filter and reinforce all the feature maps with features to produce new feature maps.
Finally, in order to get the nuclear segmentation instances, we extract the layer 1 from the new feature maps as the input to the instance head to get the predicted instance results, supervision only for labeled instances.

In addition, in order to further enhance the feature representation of nuclear, we use the proposed instance-level comparison module to make full use of the bootstrap a priori information to strengthen the nuclear features and enhance the model performance.

\begin{figure*}[tb]
\centering
  \includegraphics[width=0.99\linewidth]{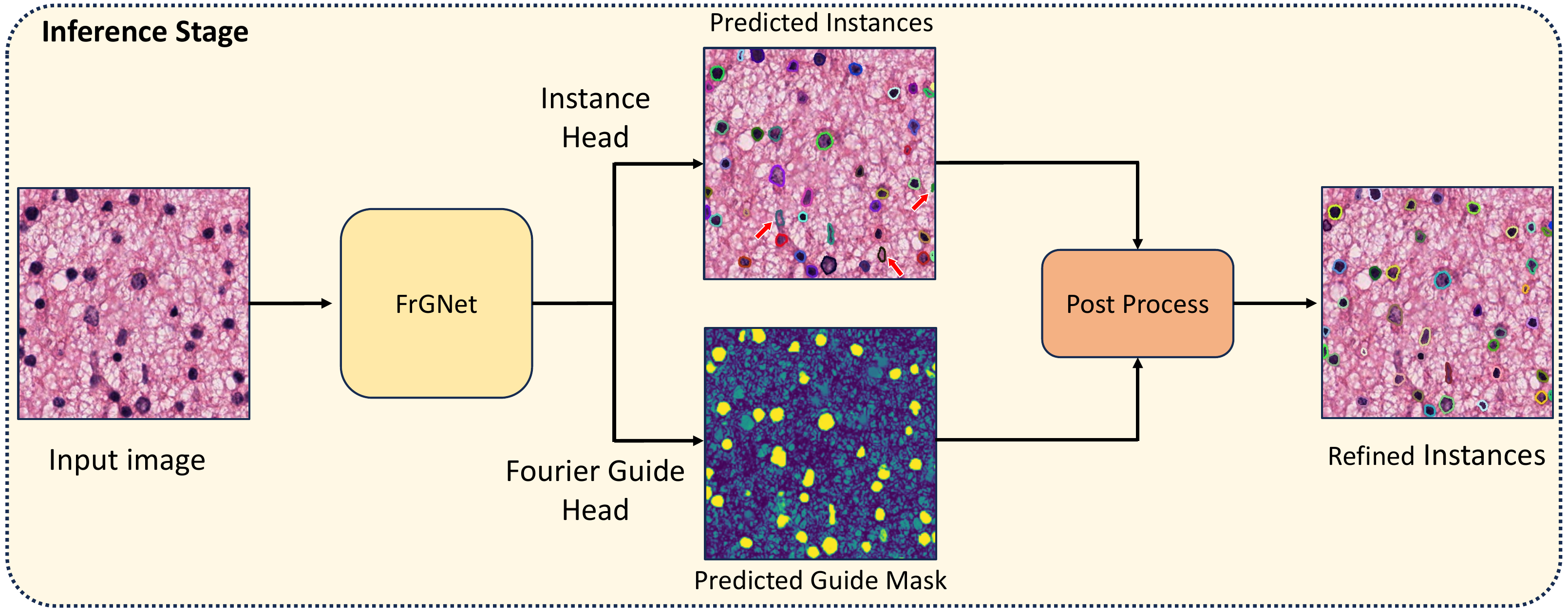}
  \caption{
  Overview of inference stage. For an input image, after processed by FrGNet, the instance head generate original predicted instances, and the fourier guide head generate predicted guide mask of input image. In post process operation, we filter out the instances in original predicted instances group according to the predicted guide mask, producing the final refined instances.
  }
\label{fig:framework_inference}
\end{figure*}

To make the manuscript self-contained, we briefly describe the necessary technical details of the original Contour proposal networks (CPN)~\citep{upschulte2022contour} first, and then introduce how we extend it with proposed fourier guide module and guide-based instance level contrastive module for nuclear instance segmentation task.

\subsection{Contour proposal networks}
Contour proposal networks (CPN) is proposed by ~\citet{upschulte2022contour}, which models nuclear instances using explicit contour expressions.
CPN consists of five parts: feature extraction backbone, classification head, contour regression head, contour refinement regression head, and post-processing.
For the input nuclear image, feature extraction is first performed using backbone (e.g., ResNet~\citep{he2015deep}).
Then, layer 2 feature map $P_2$ is extracted for instance classification and instance contour regression to obtain instance proposals.
For these instance proposals, they are sampled according to the ground truth, and only the proposals that correspond to the nuclear of the cell according to the ground truth are retained.
These retained proposals are then fine-tuned to all contours using the results generated by the contour refinement regression head, so that each predicted contour more closely matches its corresponding ground truth.
Finally, overlapping instances are removed using Non-Maximum Suppression (NMS) to get the final output instance results.

\subsection{Fourier guidance module}
As shown in Fig.~\ref{fig:teaser}, we find that the fourier transform can be used to obtain the example segmentation results of the nuclear in the pathology image of the nuclear directly, using the image characteristics of the nuclear, without any training at all.
However, such segmentation results are rough and cannot handle more complex cases.
Therefore, we inject this image characteristics information of the nuclear itself into the model training process as a kind of a priori information to guide the segmentation of the model.

\begin{figure*}[tb]
\centering
  \includegraphics[width=0.99\linewidth]{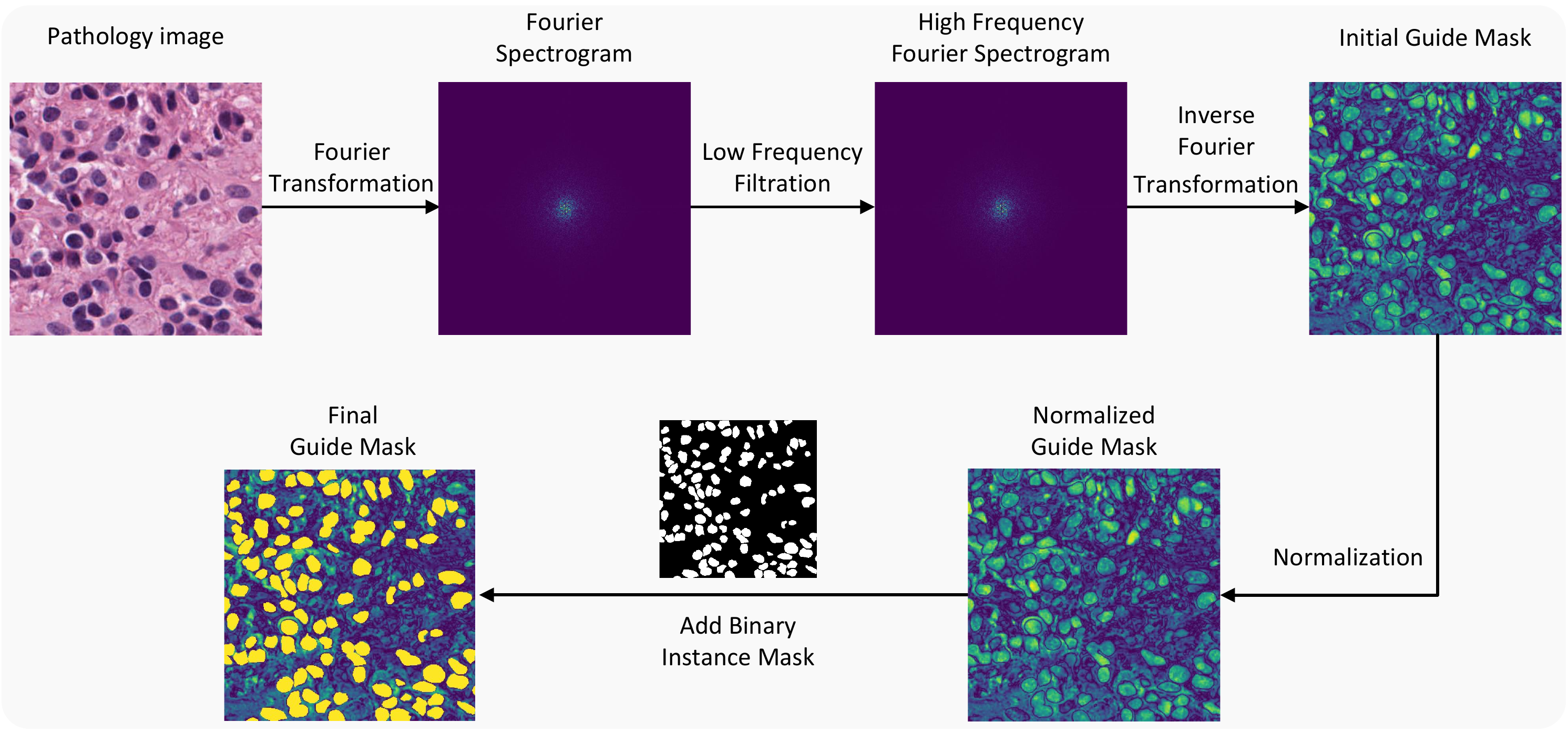}
  \caption{
  The flow of guide mask generation. For an image, we use fourier transform, low frequency filtering, inverse fourier transform to get the initial guide mask, which is then normalized and added with binary instance ground truth mask to get the final guide mask which is used to guide the model for training.
  }
\label{fig:gmask_gen}
\end{figure*}

To this end, we first design the automatic generation strategy of guide mask as shown in Fig. ~\ref{fig:gmask_gen}.
Specifically, for an input image of a nuclear pathology, we first use the fourier transformation on it to obtain the corresponding fourier spectrogram.
Second, this spectrogram is low-pass filtered using a circle with radius R (we set it as 1 during experiments) to obtain the corresponding high-frequency fourier spectrogram.
Again, for the high-frequency fourier spectrogram, we use the inverse fourier transformation to convert it back to the spatial domain image to obtain the initial guide mask.
At this time, the guide mask has more noise, in order to facilitate the processing, we carry out the normalization operation on it, so that its pixel value is normalized to between 0-1, obtaining the normalized guide mask.
Finally, we add the binary instance ground truth mask on the basis of the normalized guide mask to get the final guide mask trained with the guide model.

After obtaining the guide mask through the above process, we designed the fourier guidance module to further handle the integration of the guide mask with the model training process.
As shown in Fig.~\ref{fig:framework_train}, the input of the fourier guidance module is multi-layered feature maps.
For this feature maps, we take the layer 0 feature map $F_0$ and construct the fourier guide head, using $F_0$ as the input to get the predicted guide mask, which is supervised using $Loss_{guide\_mask}$.

Meanwhile, we utilize the predicted guide mask to filter all feature maps using the constructed GARU.
Specifically, for each feature map $F_i$, we first resize the predicted guide mask to the same size as $F_i$, and then use the GARU to bootstrap the update of $F_i$ to strengthen the features at the corresponding locations of the nuclear, and weaken the features of the non-nuclear regions to obtain the updated feature map.
For this updated feature map, we then use instance head to predict the segmentation of instances to get the final segmentation result.

\subsection{Guide-based instance level contrastive module}
In nuclear pathology images, there is a more pronounced difference between the nuclear and the background region, and a higher degree of similarity between the nuclear and the nuclear.
Therefore, the similarity between the features of different nuclear should be as similar as possible.
In our framework, the input of instance head is the feature map $F_1$ of layer 1, with size $D_1$ x $h_1$ x $w_1$.
During processing, we treat each position on $F_1$ as an instance-level feature.
Therefore, at most $h_1$*$w_1$ instance features may exist at this point.
In the actual training process, the model is able to gradually learn the nuclear instance feature representation due to the presence of ground truth.
In order to further promote the feature representation of nuclear, we utilize the guidance information provided by the guide mask to further strengthen the model's feature representation of nuclear.

Specifically, for each nuclear feature $C_i$, a vector of dimension $D_1$, we first adopt a similar projector as SimLRV2~\citep{chen2020big}, which maps $C_i$ into a hidden space of dimension $P$ to obtain nuclear feature embedding.
Subsequently, we resize the guide mask to $h_1$ x $w_1$ dimensions.
Finally, based on the resized guide mask, all the nuclear feature embedding are searched for their corresponding positive and negative samples, and supervised training is performed using the contrast loss infoNCE~\citep{oord2018representation} as a way to further enhance the model's representation of nuclear features.

\subsection{Objective functions}
To optimise our FrGNet, our loss function consists of three parts: instance, guide mask, and contrastive:
\begin{equation}
\label{eqn:loss_overall}
Loss = Loss_{instance} + Loss_{guide\_mask} + Loss_{contrastive}
\end{equation}

For instance loss, since we use contour to represents each instance, the problem of segmentation of instances is solved by transforming it into a regression of contour coordinates.
In the process of implementation we use the contour expression form in CPN and the instance loss construction, please see the original article for details.
The guide mask is generated by the fourier guide head, whose ground truth is a mask between 0 and 1. 
Therefore, we use binary cross entropy loss for supervision:
\begin{equation}
\label{eqn:loss_gmask}
Loss_{guide\_mask}(y, \hat{y}) = -\frac{1}{N} \sum_{i=1}^{N} \left[ y_i \log(\hat{y}_i) + (1 - y_i) \log(1 - \hat{y}_i) \right]
\end{equation}

Where N is the number of samples, $y_i$ is the true guide mask of the i-th sample, and $\hat{y}_i$ is the predicted obtained guide mask.

As for $Loss_{contrastive}$, we use infoNCE loss~\citep{oord2018representation} to construct, defined as below:
\begin{equation}
\label{eqn:loss_contrastive}
Loss_{contrastive} = - \frac{1}{N} \sum_{i=1}^{N} \log \frac{\exp(\text{sim}(\mathbf{z}_i, \mathbf{z}_i^+)/\tau)}{\sum_{j=1}^{N} \exp(\text{sim}(\mathbf{z}_i, \mathbf{z}_j)/\tau)}
\end{equation}

where N is the number of samples, $\mathbf{z}_i$ is the i-th instance embedding, $\mathbf{z}_i^+$ is the positive instance embedding of $\mathbf{z}_i$.
$sim()$ represents the similarity operation, we use dot product during experiments.
The $\tau$ denotes temperature coefficient, used to adjust the distribution of similarity scores, we keep it as 0.1.

\subsection{Inference stage}
The workflow of inference stage is shown in Fig.~\ref{fig:framework_inference}.
Firstly, for an input image, after processed by FrGNet, the instance head generates original predicted instances, the instances are generated in the form of contour. 
Secondly, the fourier head generates the predictions of the guide mask.
Finally, for each of the predicted instances, we perform post-processing operations in conjunction with the predicted guide mask.
Specifically, for a predicted instance, we extract the region of the guide mask where the instance is located from the predicted guide mask.
If the maximum value of the pixel values in the region's guide mask is less than a set filtering threshold, the instance is discarded, otherwise it is retained.
Such a filtering operation is performed until the last predicted instance, and the retained instances are the final output refined instances.

\section{Experiments}

\subsection{Experimental settings}

\noindent\textbf{Dataset:} 1) MoNuSeg~\citep{kumar2017dataset} contains 51 H\&E stained pathological images of size 1000×1000 from seven organs, including a total of 21,623 nuclei labeled without distinction between various categories.
In the original dataset, there are 37 for training, and 14 for testing (combining the same organ and different organ test sets);
2) CPM17~\citep{vu2019methods} contains 64 H\&E stained histopathology images with 7,570 annotated nuclear boundaries, where 32 for training and 32 for testing.
It is from the MICCAI 2017 Digital Pathology Challenge~\citep{vu2019methods} and images on two different scales: 500×500 and 600×600.
This is also a dataset for single class segmentation task.

\noindent\textbf{Evaluation metrics:} 
We use four instance-level evaluation metrics to measure the instance segmentation performance of the comparison models, which are: Aggregated Jaccard Index (AJI)~\citep{kumar2017dataset}, mean Detection Quality (DQ), mean Segmentation Quality (SQ), and mean Panoptic Quality (PQ)~\citep{kirillov2019panoptic}.
AJI computes the correlation between intersection and union pixel counts. 
It addresses the issue of over-penalization by employing ground truth with maximal intersection over union:
\begin{equation}
\text{AJI} = \frac{\sum_{i=1}^{n} |G_i \cap P_{M}^{i}|}{\sum_{i=1}^{n} |G_i \cup P_{M}^{i}| + \sum_{F \in U} |P_F|}
\end{equation}

where $n$ denotes the total number of nuclei, $P_{M}^{i}$ represents the connected regions responsible for generating mask.
$G_i$ represents the connected area where $P_{M}^{i}$ has the largest intersection with actual data. 
$U$ denotes connected regions that are not intersecting with actual data, and $P_F$ denotes element $F$ inside set $U$.
DQ, SQ and PQ defined as Eq.~\eqref{eqn:pq}.
\begin{equation}
\text{PQ} = \underbrace{\frac{\sum_{(x, y) \in TP} \text{IoU}(x, y)}{|TP|}}_{\text{segmentation quality (SQ)}} \times \underbrace{\frac{|TP|}{|TP| + \frac{1}{2}|FP| + \frac{1}{2}|FN|}}_{\text{detection quality (DQ)}}
\label{eqn:pq}
\end{equation}

where $x$ is the ground truth values, $y$ are the prediction values. 
$\text{IOU}$ denotes intersection over union and each $(x, y)$ is established as a distinct and different set. 
$TP$, $FP$, $FN$ represent true positive, false positive and false negative, respectively.

\subsection{Implementation details}
We proposed FrGNet is trained and tested using the open-source software library Pytorch 1.13.1 on 2 NVIDIA GeForce 3090 with CUDA 11.7.

\begin{figure}[tb]
\centering
  \includegraphics[width=0.99\linewidth]{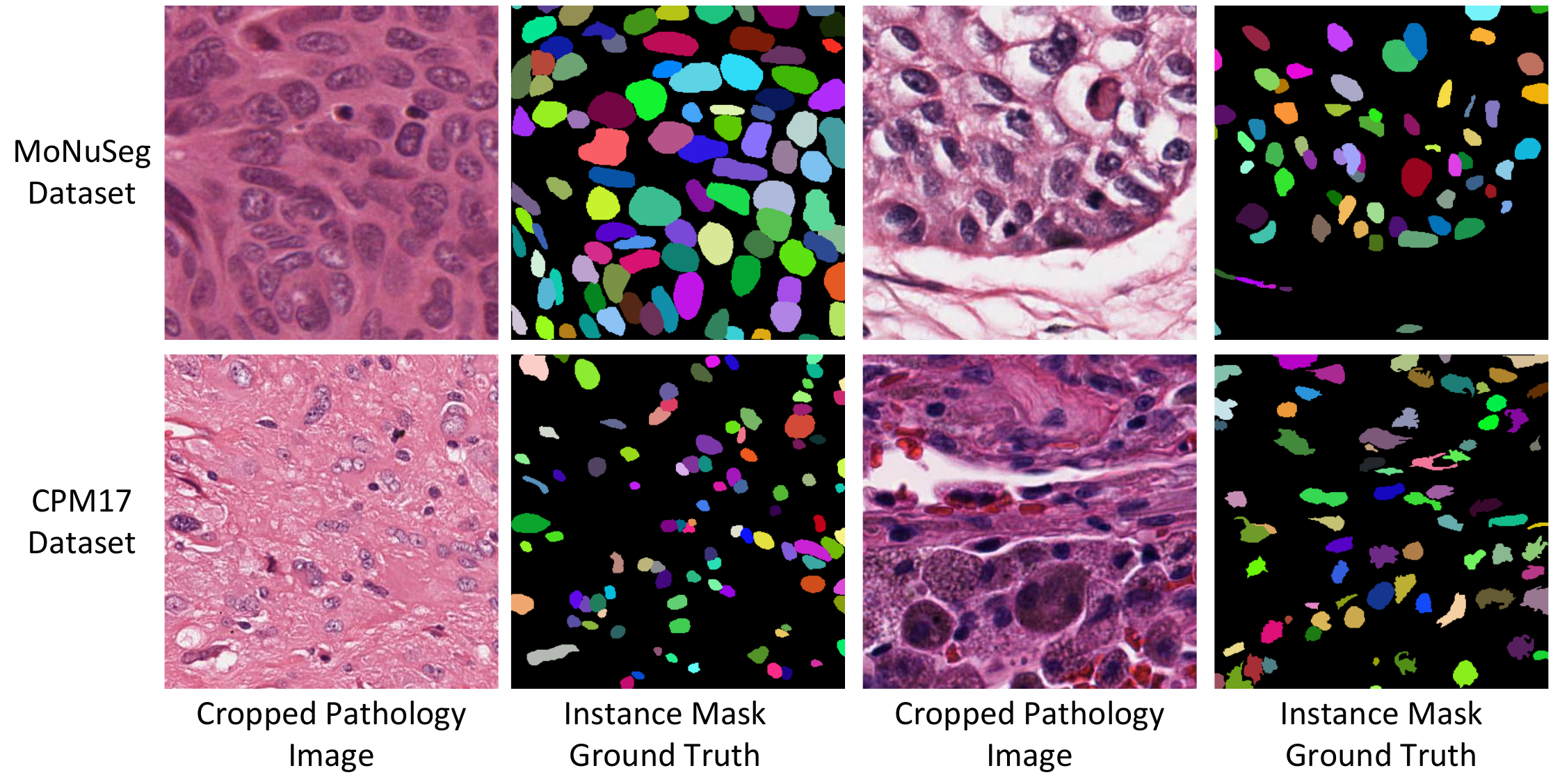}
  \caption{
  Examples of cropped histopathologic image and corresponding instance mask ground truth.
  }
\label{fig:example_cropped}
\end{figure}

Considering the size of the MoNuSeg and CPM17 datasets, we crop patches from the original histopathologic images using fixed-size boxes. 
For the cropped instances, if the area of a cropped instance is less than 10\% of the initial instance, the cropped instance is removed. 
Additionally, we remove small objects with an area of less than 20 pixels to avoid unnecessary foreground caused by incorrect pixel predictions. 
We obtain 481 patches and 416 patches from MoNuSeg and CPM17 dataset respectively. 
Examples of the cropped patches are shown in Fig.~\ref{fig:example_cropped}.

For full-supervised setting, we use all instance annotations to train proposed FrGNet model.
We train for 50 epoches with a batch size of 16 in MoNuSeg dataset.
AdamW~\citep{loshchilov2017decoupled} is adopted as the optimizer, with an initial learning rate 0.0004.
As for CPM17 dataset, We train for 30 epoches with a batch size of 16.
Adam~\citep{kingma2014adam} is adopted as the optimizer, with an initial learning rate 0.001.
Both training use warm-up and multi-step decay strategy to control the change of learning rate.
Additionally, we use replace standard batch normalization~\citep{ioffe2015batch} with synchronized batch normalization during training.

For weak-supervised setting, we first random keep 20\%, 30\%, 40\%, 60\% and 80\% instance annotations in each image to execute training.
Then, we use full annotations to test model that trained in weak-supervised setting.
Except for using hard guide, other training settings are consistent with full supervised.
During inference, we set the threshold of post process as 0.5.

\subsection{Performance comparisons in full-supervised}
We compare our weakly-supervised method with fully-
supervised methods that are trained with the completely-annotated
nuclei instances, such as U-Net~\citep{ronneberger2015u}, Mask-RCNN~\citep{he2017mask}, DCAN~\citep{chen2017dcan}, DIST~\citep{naylor2018segmentation}, Micro-Net~\citep{raza2019micro}, Full-Net~\citep{qu2019improving}, Hover-Net~\citep{graham2019hover}, PFF-Net~\citep{liu2021panoptic}, CDNet~\citep{he2021cdnet}, CPN~\citep{upschulte2022contour} and TopoSeg~\citep{he2023toposeg}.
The quantitative results are shown in Table~\ref{tab:comparison_sota}. 
As can be seen from the table, our proposed method FrGNet possesses a significant performance advantage over all current SOTAs on the Monuseg dataset, with at least 4.8\% and 2\% improvement in the PQ and AJI metrics, respectively, to achieve the new SOTA results.
Correspondingly, our method also has some advantages on the CPM17 dataset, proving the effectiveness of our method.

\begin{figure*}[tb]
\centering
  \includegraphics[width=0.99\linewidth]{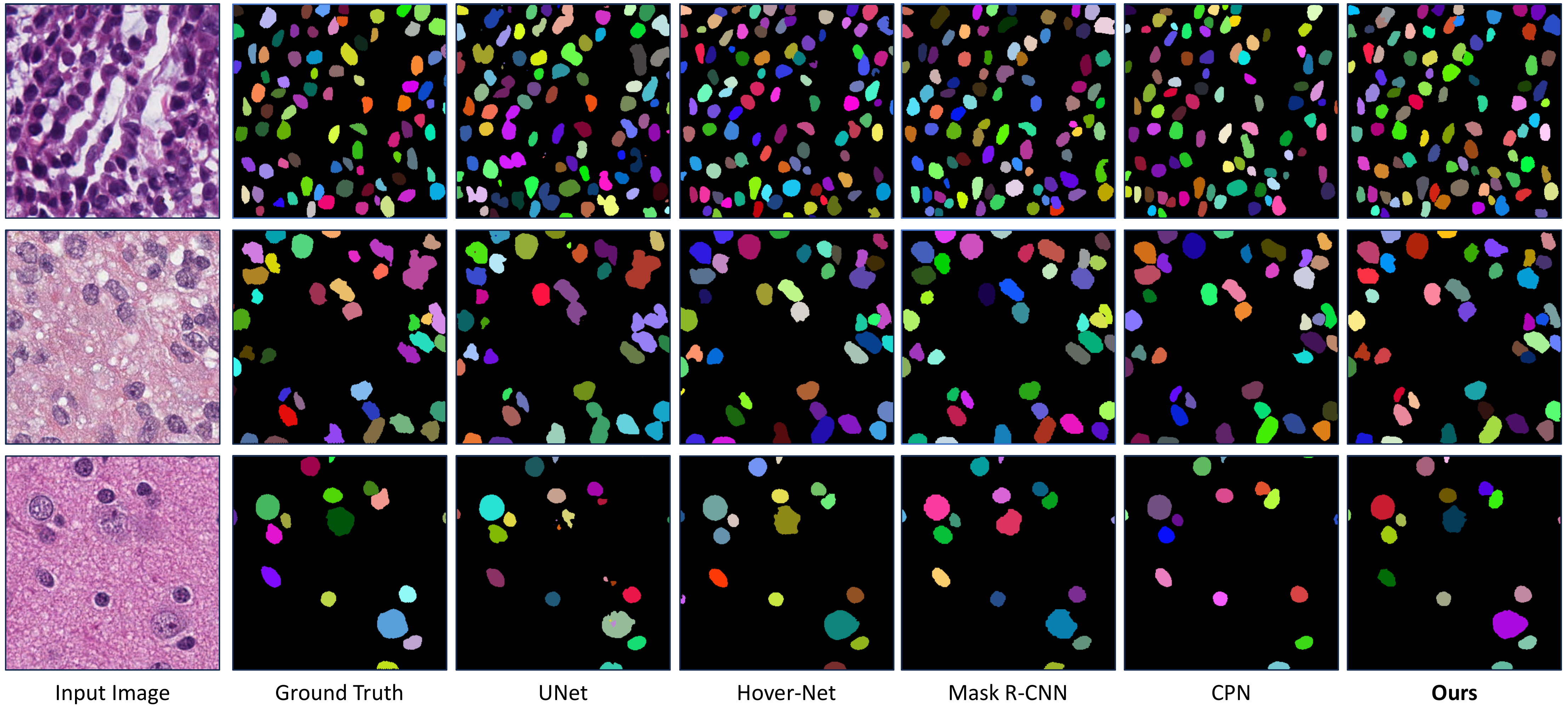}
  \caption{
  Comparisons of different full-supervised methods on MoNuSeg and CPM17 datasets.
  }
\label{fig:comp_full}
\end{figure*}

To further demonstrate the effectiveness of our framework, we performed a comparison from a visualisation point of view, and the comparison results are shown in Fig.~\ref{fig:comp_full}.
Our FrGNet not only handles the case of dense segmentation of nuclear well (first row), but also can substantially suppress the emergence of false positive instances when the distribution of nuclear is more sparse (the second and the third rows), which is attributed to the design of our Fourier-guided model architecture.

\begin{table}[tb]
\caption{Performance comparisons of different full-supervised methods on MoNuSeg and CPM17 datasets.}
\label{tab:comparison_sota}
\begingroup 
\renewcommand{\arraystretch}{1.7}
\resizebox{1.0\linewidth}{!}{
\begin{tabular}{c|cc|cc}
\hline
          \multirow{2}{*}{Method} & \multicolumn{2}{c|}{MoNuSeg}  & \multicolumn{2}{c}{CPM17}    \\ \cline{2-5}
          & \multicolumn{1}{c}{PQ $\uparrow$} & AJI $\uparrow$ & \multicolumn{1}{c}{PQ $\uparrow$} & AJI $\uparrow$ \\  \hline
U-Net~\citep{ronneberger2015u}     & \multicolumn{1}{c}{$58.1$}   & $57.9$  & \multicolumn{1}{c}{$62.5$}   & $66.6$  \\ 
Mask-RCNN~\citep{he2017mask} & \multicolumn{1}{c}{$50.9$}   & $54.6$ & \multicolumn{1}{c}{$67.4$}   &  $68.4$   \\ 
DCAN~\citep{chen2017dcan}      & \multicolumn{1}{c}{$49.2$}   &  $52.5$   & \multicolumn{1}{c}{$54.5$}   &   $56.1$  \\ 
DIST~\citep{naylor2018segmentation}      & \multicolumn{1}{c}{$44.3$}   &  $55.9$   & \multicolumn{1}{c}{$50.4$}   &   $61.6$  \\ 
Micro-Net~\citep{raza2019micro} & \multicolumn{1}{c}{$51.9$}   &  $56.0$   & \multicolumn{1}{c}{$66.1$}   &  $66.8$   \\ 
Full-Net~\citep{qu2019improving}  & \multicolumn{1}{c}{-}   &   $60.4$  & \multicolumn{1}{c}{-}   &  $66.1$   \\ 
Hover-Net~\citep{graham2019hover} & \multicolumn{1}{c}{$59.7$}   &  $61.8$   & \multicolumn{1}{c}{$69.7$}   &   $70.5$  \\ 
PFF-Net~\citep{liu2021panoptic}   & \multicolumn{1}{c}{$58.7$}   &  $61.1$  & \multicolumn{1}{c}{-}   &   -  \\ 
CDNet~\citep{he2021cdnet}     & \multicolumn{1}{c}{-}   &  $63.7$   & \multicolumn{1}{c}{-}   &  $73.3$   \\ 
CPN~\citep{upschulte2022contour}     & \multicolumn{1}{c}{$62.7$}   &  $61.5$   & \multicolumn{1}{c}{68.9}   &  $69.1$   \\ 
TopoSeg~\citep{he2023toposeg}   & \multicolumn{1}{c}{$62.5$}   &  $64.3$   & \multicolumn{1}{c}{$70.5$}   &  $75.6$   \\  \hline
\textbf{FrGNet (Ours)}      & \multicolumn{1}{c}{\textbf{65.5}}   &   \textbf{65.6}  & \multicolumn{1}{c}{\textbf{70.7}}   &   71.2  \\ \hline
\end{tabular}
}
\endgroup
\end{table}

\subsection{Performance comparisons in weak-supervised}
\begin{figure*}[tb]
\centering
  \includegraphics[width=0.99\linewidth]{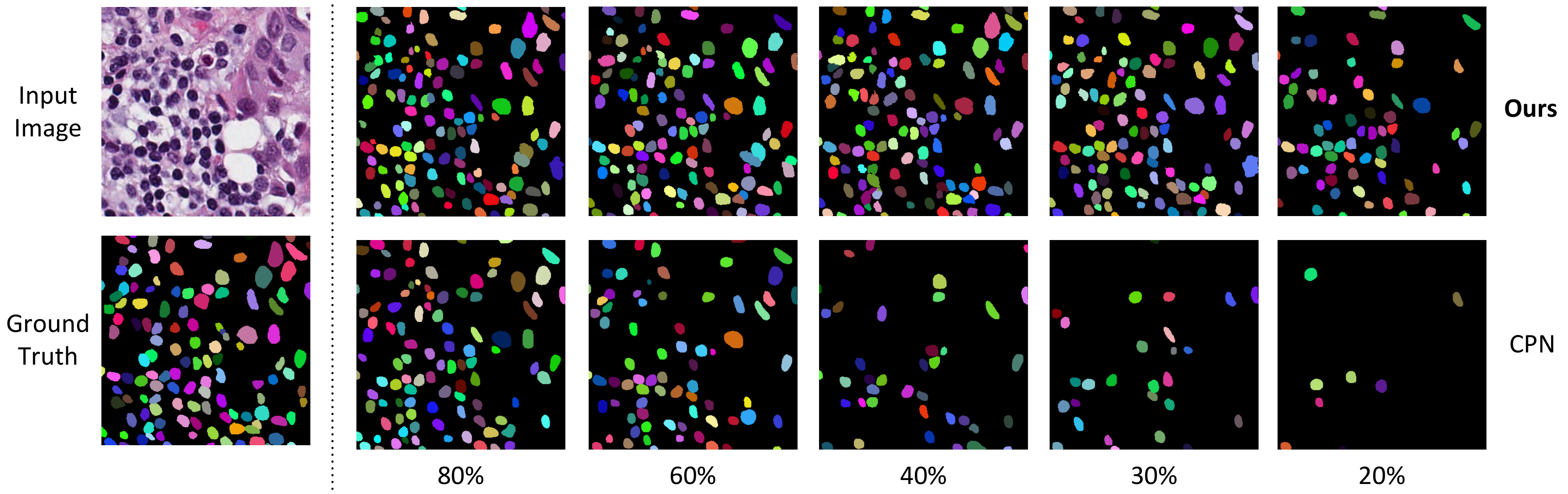}
  \caption{
  Comparisons of different weak-supervised methods on MoNuSeg and CPM17 datasets.
  }
\label{fig:comp_weak}
\end{figure*}

The quantitative results of the method proposed in this paper and the existing methods under weakly supervised design are shown in Table 2.
Under weak supervision, we randomly remove 20\%, 40\%, 60\%, 70\%, and 80\% of the instance annotations in each image, followed by training, and testing with the same test set as full supervision.
As can be seen from the table, the quantitative metrics of baseline show a linear decrease as the instance annotations are reduced, while our method always maintains a performance that is not too different from that of full supervision.
Structurally, the shape of the nucleus is basically similar, so the model can learn the shape characteristics of the nucleus with little data.
The localisation of nucleus instances can be done by our Fourier guidance module, and thus the method proposed in this paper still has good performance even with a rather small amount of annotations.
The visualization results are shown in Fig.~\ref{fig:comp_weak}.

\begin{table}[tb]
\caption{Performance comparisons of different weak-supervised methods on MoNuSeg and CPM17 datasets.}
\label{tab:comparison_weak}
\begingroup 
\renewcommand{\arraystretch}{1.35}
\resizebox{1.0\linewidth}{!}{
\begin{tabular}{c|c|cc|cc}
\hline
    \multirow{2}{*}{\makecell{\% keep annotation \\of each image}} & \multirow{2}{*}{Method} & \multicolumn{2}{c|}{MoNuSeg} & \multicolumn{2}{c}{CPM17} \\                                                      
    \cline{3-6}
 &  & \multicolumn{1}{c}{PQ $\uparrow$} & AJI $\uparrow$ & \multicolumn{1}{c}{PQ $\uparrow$} & AJI $\uparrow$ \\ 
 \hline
 \multirow{2}{*}{20\%}   &   CPN   & \multicolumn{1}{c}{13.6}   &  8.6   & \multicolumn{1}{c}{38.4}   &   28.9  \\ 
 &   \textbf{FrGNet (Ours)}   & \multicolumn{1}{c}{\textbf{49.2}}   & \textbf{40.5}   & \multicolumn{1}{c}{\textbf{52.5}}   &   \textbf{43.3}  \\ 
\hline
 \multirow{2}{*}{30\%}   &   CPN   & \multicolumn{1}{c}{21.2}   &  13.5   & \multicolumn{1}{c}{39.8}   &   29.3 \\ 
 &   \textbf{FrGNet (Ours)}   & \multicolumn{1}{c}{\textbf{62.6}}   & \textbf{61.9}   & \multicolumn{1}{c}{\textbf{64.8}}   & \textbf{61.6}     \\  
\hline
 \multirow{2}{*}{40\%}   &   CPN   & \multicolumn{1}{c}{28.5}   &   19.2  & \multicolumn{1}{c}{52.0}   &  42.5   \\ 
 &   \textbf{FrGNet (Ours)}   & \multicolumn{1}{c}{\textbf{63.8}}   &  \textbf{62.7}   & \multicolumn{1}{c}{\textbf{66.9}}   &  \textbf{64.7}   \\ 
\hline
 \multirow{2}{*}{60\%}   &   CPN   & \multicolumn{1}{c}{46.8}   &  37.8   & \multicolumn{1}{c}{63.6}   &  59.2   \\ 
 &   \textbf{FrGNet (Ours)}   & \multicolumn{1}{c}{\textbf{63.9}}   &  \textbf{63.2}   & \multicolumn{1}{c}{\textbf{69.6}}   &  \textbf{68.4}   \\ 

\hline
 \multirow{2}{*}{80\%}   &   CPN   & \multicolumn{1}{c}{56.8}   &  51.2  & \multicolumn{1}{c}{67.5}   &  65.2   \\ 
 &   \textbf{FrGNet (Ours)}   & \multicolumn{1}{c}{\textbf{63.2}}   &   \textbf{62.6}  & \multicolumn{1}{c}{\textbf{70.1}}   &   \textbf{70.1}  \\ 
\hline
 \multirow{2}{*}{100\%}   &   CPN   & \multicolumn{1}{c}{62.7}   &  61.5   & \multicolumn{1}{c}{68.9}   &  69.1  \\ 
 &   \textbf{FrGNet (Ours)}   & \multicolumn{1}{c}{\textbf{65.5}}   &   \textbf{65.6}  & \multicolumn{1}{c}{\textbf{70.7}}   &   \textbf{71.2}  \\ 
\hline
\end{tabular}
}
\endgroup
\end{table}

\subsection{Ablation study}
We performed ablation studies on the CPM17 and MoNuSeg datasets to validate the efficacy of the proposed FrGNet method.
All the following experiments were conducted under the setting of fully supervised tasks.

\noindent\textbf{Effectiveness of FG and GILC modules}. Our proposed FG module provides guiding information about the location of the nucleus for the whole framework, and the GILC module enhances the feature representation of the nucleus using instance-level information.
Table~\ref{tab:comparison_ablation1} shows the changes in the performance metrics of the model when the FG and GILC modules are used or not.
From the table, it can be seen that when both FG and GILC modules are not used (\#0), the model does not introduce Fourier bootstrap information at this time, and thus has the worst performance.
After the FG module is used (\#1), the model's performance is improved to some extent thanks to the introduction of the a priori bootstrap information.
And when the FG and GILC modules are used at the same time (\#2), the two modules promote each other and work together to provide the model with optimisation information, which promotes the model to further learn further about the feature representation of the nucleus as well as the distribution of the nuclear location, in which case the model performance is highest.
\begin{table}[tb]
\caption{Ablation studies of FG and GILC modules.}
\label{tab:comparison_ablation1}
\begingroup 
\renewcommand{\arraystretch}{1.9}
\resizebox{1\linewidth}{!}{
\begin{tabular}{c|cc|cccc|cccc}
\hline
   \multirow{2}{*}{\#} &  \multirow{2}{*}{\makecell{FG\\ Module}} & \multirow{2}{*}{\makecell{GILC\\Module}} & \multicolumn{4}{c|}{MoNuSeg} & \multicolumn{4}{c}{CPM17} \\                                                      
    \cline{4-11}
 &  &  & DQ $\uparrow$ & SQ $\uparrow$ & PQ $\uparrow$ & AJI $\uparrow$ & DQ $\uparrow$ & SQ $\uparrow$ & PQ $\uparrow$ & AJI $\uparrow$ \\ 
 \hline
0 &  \XSolidBrush   &   \XSolidBrush    &  82.2  &  76.2  &  62.7  &  61.5   &  85.9  &  80.1  &  68.9  &  69.1   \\ 
1 &  \Checkmark   &   \XSolidBrush   &  82.7  &  75.9  &  62.8  &  63.0  &  86.9  &  79.9  &  69.5  &  70.5   \\ 
2 &  \Checkmark   & \Checkmark   &  \textbf{84.8}  &  \textbf{77.2}  &  \textbf{65.5}  &  \textbf{65.6}  &  \textbf{87.5} &  \textbf{80.6} &  \textbf{70.7}  &  \textbf{71.2}    \\ 
\hline
\end{tabular}
}
\endgroup
\end{table}


\noindent\textbf{Type of guide mask}. By generating a fourier guidance mask (as shown in Fig.~\ref{fig:gmask_gen}) is not a binary mask, but a floating-point form of mask.
In concert with the commonly used and effective label smoothing trick~\citep{muller2019does}, we further explored the impact of the supervised form of the guide mask on model performance.
Specifically, the guide mask generated by Fig.~\ref{fig:gmask_gen} is the soft form of guide mask, and the hard guide mask is obtained by binarising the soft guide mask.
After obtaining the soft guide mask and the hard guide mask, we explored the effect of the type of guide mask on the performance of the model, and the results are shown in Table ~\ref{tab:comparison_ablation2}.
From this table, it can be seen that when no guide mask is used, the model performs poorly (\#0) due to the missing guide information.
When the guide mask is used, the model performance is improved (\#1\&2).
At the same time, the use of soft form of guide mask has the greatest improvement in the performance of the model, this is because soft form of guide mask not only brings a priori guidance information, but also brings feature information to the model that cannot be brought by the hard form, and these feature information makes the model more robust.
\begin{table}[tb]
\caption{Ablation studies on the type of guide mask.}
\label{tab:comparison_ablation2}
\begingroup 
\renewcommand{\arraystretch}{1.9}
\resizebox{1\linewidth}{!}{
\begin{tabular}{c|cc|cccc|cccc}
\hline
   \multirow{2}{*}{\#} &  \multirow{2}{*}{\makecell{Hard\\ Guide}} & \multirow{2}{*}{\makecell{Soft\\Guide}} & \multicolumn{4}{c|}{MoNuSeg} & \multicolumn{4}{c}{CPM17} \\                                                      
    \cline{4-11}
 &  &  & DQ $\uparrow$ & SQ $\uparrow$ & PQ $\uparrow$ & AJI $\uparrow$ & DQ $\uparrow$ & SQ $\uparrow$ & PQ $\uparrow$ & AJI $\uparrow$ \\ 
 \hline
0 &  \XSolidBrush   &   \XSolidBrush    &  80.7  &  \textbf{77.8}  &  62.9  &  61.7  &  86.8  &  80.2  &  69.7  &  70.3   \\ 
1 &  \Checkmark   &   \XSolidBrush   &  84.5  &  77.4  &  65.5  &  65.1   &  87.0  &  80.2  &  69.9  &   70.2   \\ 
2 &  \XSolidBrush   & \Checkmark   &  \textbf{84.8}  &  77.2  &  \textbf{65.5}  &  \textbf{65.6}  &  \textbf{87.5} &  \textbf{80.6} &  \textbf{70.7}  &  \textbf{71.2}    \\ 
\hline
\end{tabular}
}
\endgroup
\end{table}


\subsection{Generalization}
\begin{figure*}[tb]
\centering
  \includegraphics[width=0.93\linewidth]{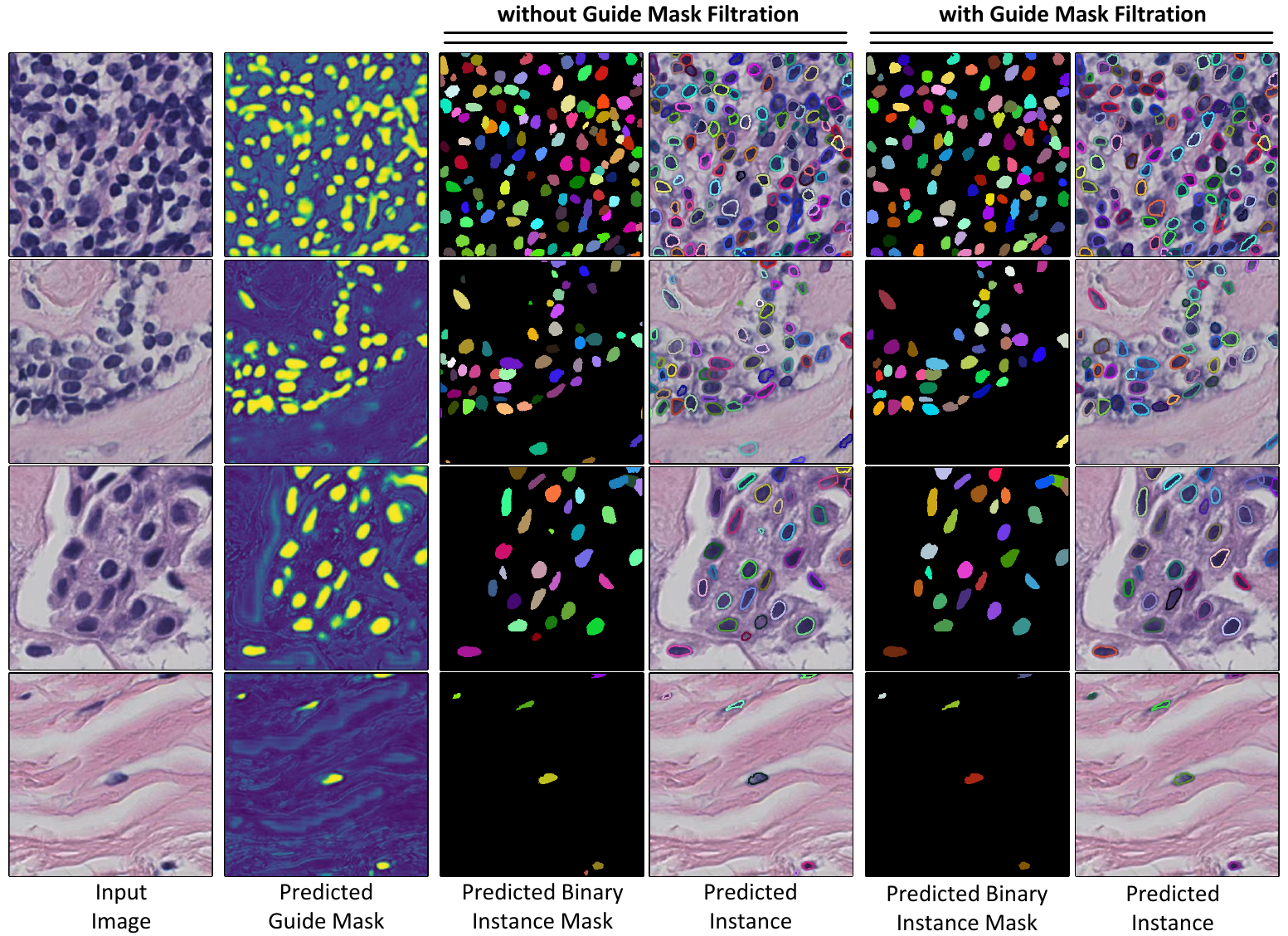}
  \caption{
  Generalization results on private dataset.
  }
\label{fig:genera}
\end{figure*}

As shown in Table~\ref{tab:comparison_weak}, our proposed method is able to keep stable performance when data annotations decreasing.
In order to further explore the generalization ability of FrGNet, we use a private dataset.
This dataset consist of 3100 histopathologic images, without any instance annotations.
Specially, we initial our FrGNet model with the model weight that training on the MoNuSeg dataset in full-supervised setting.
Then, we fine-tune this model 5 epoches, with freeze backbone and other heads except \textbf{Fourier Guide Head}.
Finally, we use this fine-tuned FrGNet to do inference on this private dataset.
The visualization results are shown in Fig.~\ref{fig:genera}.
\textbf{Without any annotations}, our proposed FrGNet can generate more complete nuclear segmentation.
On the one hand, our algorithm can segment the nuclear nicely no matter they are densely or sparsely distributed.
On the other hand, we can make full use of the guide mask to filter the segmentation results, which effectively mitigates the problem of false positives, and at the same time, the guide mask shows the possible locations of the nuclear in the model output, which strengthens the interpretability of the model.

\section{Limitations and discussion}

In this work, we present a framework for segmenting nuclear instances based on fourier guidance.
The framework incorporates the feature information of the nuclear image itself, and injects the a priori information of the nuclear location into the model by using a constructed fourier guidance (FG) module to guide the model for training.
To further improve the model performance, we propose the guide-based instance level contrastive (GILC) module, which provides instance-level feature guidance for the model and enhances the feature representation of the nuclear instances.
Through comparison and ablation experiments, the effectiveness of our framework and the proposed modules are demonstrated to reach a new SOTA in performance.
Meanwhile, it is demonstrated through experiments that thanks to the construction of fourier guidance information, our framework is able to solve both fully supervised and weakly supervised nuclear instance segmentation problems.
Besides, our model possesses good generalisation.
By working on a private nuclear pathology dataset without any supervised annotations, our model is still able to perform the task of instance segmentation of nuclear well in data that has not been seen at the time of training.

Nuclear in pathology images are generally difficult to annotate.
How to construct a completely unsupervised framework for segmenting nuclear instances is a task of its own interest, and solving the problem will largely contribute to the development of nuclear segmentation in the field of pathology.
The framework proposed in this paper, although possessing a certain degree of generalisation, still needs to rely on a pre-trained model on fully supervised dataset, and is unable to accomplish the task of unsupervised nuclear instance segmentation in the true sense.
Recently the field of unsupervised segmentation has gained some development~\citep{sun2024clip, niu2024unsupervised}, which is of some guidance for unsupervised nuclear instance segmentation.







\bibliographystyle{model2-names.bst}\biboptions{authoryear}
\bibliography{refs}

\begin{thebibliography}{52}
\expandafter\ifx\csname natexlab\endcsname\relax\def\natexlab#1{#1}\fi
\providecommand{\url}[1]{\texttt{#1}}
\providecommand{\href}[2]{#2}
\providecommand{\path}[1]{#1}
\providecommand{\DOIprefix}{doi:}
\providecommand{\ArXivprefix}{arXiv:}
\providecommand{\URLprefix}{URL: }
\providecommand{\Pubmedprefix}{pmid:}
\providecommand{\doi}[1]{\href{http://dx.doi.org/#1}{\path{#1}}}
\providecommand{\Pubmed}[1]{\href{pmid:#1}{\path{#1}}}
\providecommand{\bibinfo}[2]{#2}
\ifx\xfnm\relax \def\xfnm[#1]{\unskip,\space#1}\fi
\bibitem[{Abbas(2014)}]{abbas2014occluded}
\bibinfo{author}{Abbas, N.}, \bibinfo{year}{2014}.
\newblock \bibinfo{title}{Occluded red blood cells splitting via boundaries analysis and lines drawing in microscopic thin blood smear digital images}.
\newblock \bibinfo{journal}{VFAST Transactions on Software Engineering} \bibinfo{volume}{2}, \bibinfo{pages}{100--107}.
\bibitem[{Bearman et~al.(2016)Bearman, Russakovsky, Ferrari and Fei-Fei}]{bearman2016s}
\bibinfo{author}{Bearman, A.}, \bibinfo{author}{Russakovsky, O.}, \bibinfo{author}{Ferrari, V.}, \bibinfo{author}{Fei-Fei, L.}, \bibinfo{year}{2016}.
\newblock \bibinfo{title}{What’s the point: Semantic segmentation with point supervision}, in: \bibinfo{booktitle}{European conference on computer vision}, \bibinfo{organization}{Springer}. pp. \bibinfo{pages}{549--565}.
\bibitem[{Chaitanya et~al.(2020)Chaitanya, Erdil, Karani and Konukoglu}]{chaitanya2020contrastive}
\bibinfo{author}{Chaitanya, K.}, \bibinfo{author}{Erdil, E.}, \bibinfo{author}{Karani, N.}, \bibinfo{author}{Konukoglu, E.}, \bibinfo{year}{2020}.
\newblock \bibinfo{title}{Contrastive learning of global and local features for medical image segmentation with limited annotations}.
\newblock \bibinfo{journal}{Advances in neural information processing systems} \bibinfo{volume}{33}, \bibinfo{pages}{12546--12558}.
\bibitem[{Chen et~al.(2017)Chen, Qi, Yu, Dou, Qin and Heng}]{chen2017dcan}
\bibinfo{author}{Chen, H.}, \bibinfo{author}{Qi, X.}, \bibinfo{author}{Yu, L.}, \bibinfo{author}{Dou, Q.}, \bibinfo{author}{Qin, J.}, \bibinfo{author}{Heng, P.A.}, \bibinfo{year}{2017}.
\newblock \bibinfo{title}{Dcan: Deep contour-aware networks for object instance segmentation from histology images}.
\newblock \bibinfo{journal}{Medical image analysis} \bibinfo{volume}{36}, \bibinfo{pages}{135--146}.
\bibitem[{Chen et~al.(2023)Chen, Ding, Liu, Cheng and Tao}]{chen2023cpp}
\bibinfo{author}{Chen, S.}, \bibinfo{author}{Ding, C.}, \bibinfo{author}{Liu, M.}, \bibinfo{author}{Cheng, J.}, \bibinfo{author}{Tao, D.}, \bibinfo{year}{2023}.
\newblock \bibinfo{title}{Cpp-net: Context-aware polygon proposal network for nucleus segmentation}.
\newblock \bibinfo{journal}{IEEE Transactions on Image Processing} \bibinfo{volume}{32}, \bibinfo{pages}{980--994}.
\bibitem[{Chen et~al.(2020a)Chen, Kornblith, Norouzi and Hinton}]{chen2020simple}
\bibinfo{author}{Chen, T.}, \bibinfo{author}{Kornblith, S.}, \bibinfo{author}{Norouzi, M.}, \bibinfo{author}{Hinton, G.}, \bibinfo{year}{2020}a.
\newblock \bibinfo{title}{A simple framework for contrastive learning of visual representations}, in: \bibinfo{booktitle}{International conference on machine learning}, \bibinfo{organization}{PMLR}. pp. \bibinfo{pages}{1597--1607}.
\bibitem[{Chen et~al.(2020b)Chen, Kornblith, Swersky, Norouzi and Hinton}]{chen2020big}
\bibinfo{author}{Chen, T.}, \bibinfo{author}{Kornblith, S.}, \bibinfo{author}{Swersky, K.}, \bibinfo{author}{Norouzi, M.}, \bibinfo{author}{Hinton, G.E.}, \bibinfo{year}{2020}b.
\newblock \bibinfo{title}{Big self-supervised models are strong semi-supervised learners}.
\newblock \bibinfo{journal}{Advances in neural information processing systems} \bibinfo{volume}{33}, \bibinfo{pages}{22243--22255}.
\bibitem[{Chen et~al.(2020c)Chen, Fan, Girshick and He}]{chen2020improved}
\bibinfo{author}{Chen, X.}, \bibinfo{author}{Fan, H.}, \bibinfo{author}{Girshick, R.}, \bibinfo{author}{He, K.}, \bibinfo{year}{2020}c.
\newblock \bibinfo{title}{Improved baselines with momentum contrastive learning}.
\newblock \bibinfo{journal}{arXiv preprint arXiv:2003.04297} .
\bibitem[{Cheng et~al.(2008)Cheng, Rajapakse et~al.}]{cheng2008segmentation}
\bibinfo{author}{Cheng, J.}, \bibinfo{author}{Rajapakse, J.C.}, et~al., \bibinfo{year}{2008}.
\newblock \bibinfo{title}{Segmentation of clustered nuclei with shape markers and marking function}.
\newblock \bibinfo{journal}{IEEE transactions on Biomedical Engineering} \bibinfo{volume}{56}, \bibinfo{pages}{741--748}.
\bibitem[{Dai et~al.(2015)Dai, He and Sun}]{dai2015boxsup}
\bibinfo{author}{Dai, J.}, \bibinfo{author}{He, K.}, \bibinfo{author}{Sun, J.}, \bibinfo{year}{2015}.
\newblock \bibinfo{title}{Boxsup: Exploiting bounding boxes to supervise convolutional networks for semantic segmentation}, in: \bibinfo{booktitle}{Proceedings of the IEEE international conference on computer vision}, pp. \bibinfo{pages}{1635--1643}.
\bibitem[{Graham et~al.(2019)Graham, Vu, Raza, Azam, Tsang, Kwak and Rajpoot}]{graham2019hover}
\bibinfo{author}{Graham, S.}, \bibinfo{author}{Vu, Q.D.}, \bibinfo{author}{Raza, S.E.A.}, \bibinfo{author}{Azam, A.}, \bibinfo{author}{Tsang, Y.W.}, \bibinfo{author}{Kwak, J.T.}, \bibinfo{author}{Rajpoot, N.}, \bibinfo{year}{2019}.
\newblock \bibinfo{title}{Hover-net: Simultaneous segmentation and classification of nuclei in multi-tissue histology images}.
\newblock \bibinfo{journal}{Medical image analysis} \bibinfo{volume}{58}, \bibinfo{pages}{101563}.
\bibitem[{Grill et~al.(2020)Grill, Strub, Altch{\'e}, Tallec, Richemond, Buchatskaya, Doersch, Avila~Pires, Guo, Gheshlaghi~Azar et~al.}]{grill2020bootstrap}
\bibinfo{author}{Grill, J.B.}, \bibinfo{author}{Strub, F.}, \bibinfo{author}{Altch{\'e}, F.}, \bibinfo{author}{Tallec, C.}, \bibinfo{author}{Richemond, P.}, \bibinfo{author}{Buchatskaya, E.}, \bibinfo{author}{Doersch, C.}, \bibinfo{author}{Avila~Pires, B.}, \bibinfo{author}{Guo, Z.}, \bibinfo{author}{Gheshlaghi~Azar, M.}, et~al., \bibinfo{year}{2020}.
\newblock \bibinfo{title}{Bootstrap your own latent-a new approach to self-supervised learning}.
\newblock \bibinfo{journal}{Advances in neural information processing systems} \bibinfo{volume}{33}, \bibinfo{pages}{21271--21284}.
\bibitem[{He et~al.(2021a)He, Huang, Ding, Song, Wang, Ren, Wei, Gao and Chen}]{he2021cdnet}
\bibinfo{author}{He, H.}, \bibinfo{author}{Huang, Z.}, \bibinfo{author}{Ding, Y.}, \bibinfo{author}{Song, G.}, \bibinfo{author}{Wang, L.}, \bibinfo{author}{Ren, Q.}, \bibinfo{author}{Wei, P.}, \bibinfo{author}{Gao, Z.}, \bibinfo{author}{Chen, J.}, \bibinfo{year}{2021}a.
\newblock \bibinfo{title}{Cdnet: Centripetal direction network for nuclear instance segmentation}, in: \bibinfo{booktitle}{Proceedings of the IEEE/CVF International Conference on Computer Vision}, pp. \bibinfo{pages}{4026--4035}.
\bibitem[{He et~al.(2023a)He, Wang, Wei, Xu, Ji, Liu and Chen}]{he2023toposeg}
\bibinfo{author}{He, H.}, \bibinfo{author}{Wang, J.}, \bibinfo{author}{Wei, P.}, \bibinfo{author}{Xu, F.}, \bibinfo{author}{Ji, X.}, \bibinfo{author}{Liu, C.}, \bibinfo{author}{Chen, J.}, \bibinfo{year}{2023}a.
\newblock \bibinfo{title}{Toposeg: Topology-aware nuclear instance segmentation}, in: \bibinfo{booktitle}{Proceedings of the IEEE/CVF International Conference on Computer Vision}, pp. \bibinfo{pages}{21307--21316}.
\bibitem[{He et~al.(2021b)He, Zhang, Chen, Geng, Chen, Liang, Lu, Wu and Xu}]{he2021hybrid}
\bibinfo{author}{He, H.}, \bibinfo{author}{Zhang, C.}, \bibinfo{author}{Chen, J.}, \bibinfo{author}{Geng, R.}, \bibinfo{author}{Chen, L.}, \bibinfo{author}{Liang, Y.}, \bibinfo{author}{Lu, Y.}, \bibinfo{author}{Wu, J.}, \bibinfo{author}{Xu, Y.}, \bibinfo{year}{2021}b.
\newblock \bibinfo{title}{A hybrid-attention nested unet for nuclear segmentation in histopathological images}.
\newblock \bibinfo{journal}{Frontiers in Molecular Biosciences} \bibinfo{volume}{8}, \bibinfo{pages}{614174}.
\bibitem[{He et~al.(2017)He, Gkioxari, Doll{\'a}r and Girshick}]{he2017mask}
\bibinfo{author}{He, K.}, \bibinfo{author}{Gkioxari, G.}, \bibinfo{author}{Doll{\'a}r, P.}, \bibinfo{author}{Girshick, R.}, \bibinfo{year}{2017}.
\newblock \bibinfo{title}{Mask r-cnn}, in: \bibinfo{booktitle}{Proceedings of the IEEE international conference on computer vision}, pp. \bibinfo{pages}{2961--2969}.
\bibitem[{He et~al.(2015)He, Zhang, Ren et~al.}]{he2015deep}
\bibinfo{author}{He, K.}, \bibinfo{author}{Zhang, X.}, \bibinfo{author}{Ren, S.}, et~al., \bibinfo{year}{2015}.
\newblock \bibinfo{title}{Deep residual learning}.
\newblock \bibinfo{journal}{Image Recognition} .
\bibitem[{He et~al.(2023b)He, Unberath, Ke and Shen}]{he2023transnuseg}
\bibinfo{author}{He, Z.}, \bibinfo{author}{Unberath, M.}, \bibinfo{author}{Ke, J.}, \bibinfo{author}{Shen, Y.}, \bibinfo{year}{2023}b.
\newblock \bibinfo{title}{Transnuseg: A lightweight multi-task transformer for nuclei segmentation}, in: \bibinfo{booktitle}{International Conference on Medical Image Computing and Computer-Assisted Intervention}, \bibinfo{organization}{Springer}. pp. \bibinfo{pages}{206--215}.
\bibitem[{Ioffe and Szegedy(2015)}]{ioffe2015batch}
\bibinfo{author}{Ioffe, S.}, \bibinfo{author}{Szegedy, C.}, \bibinfo{year}{2015}.
\newblock \bibinfo{title}{Batch normalization: Accelerating deep network training by reducing internal covariate shift}, in: \bibinfo{booktitle}{International conference on machine learning}, \bibinfo{organization}{pmlr}. pp. \bibinfo{pages}{448--456}.
\bibitem[{Kervadec et~al.(2019)Kervadec, Dolz, Tang, Granger, Boykov and Ayed}]{kervadec2019constrained}
\bibinfo{author}{Kervadec, H.}, \bibinfo{author}{Dolz, J.}, \bibinfo{author}{Tang, M.}, \bibinfo{author}{Granger, E.}, \bibinfo{author}{Boykov, Y.}, \bibinfo{author}{Ayed, I.B.}, \bibinfo{year}{2019}.
\newblock \bibinfo{title}{Constrained-cnn losses for weakly supervised segmentation}.
\newblock \bibinfo{journal}{Medical image analysis} \bibinfo{volume}{54}, \bibinfo{pages}{88--99}.
\bibitem[{Kingma and Ba(2014)}]{kingma2014adam}
\bibinfo{author}{Kingma, D.P.}, \bibinfo{author}{Ba, J.}, \bibinfo{year}{2014}.
\newblock \bibinfo{title}{Adam: A method for stochastic optimization}.
\newblock \bibinfo{journal}{arXiv preprint arXiv:1412.6980} .
\bibitem[{Kirillov et~al.(2019)Kirillov, He, Girshick, Rother and Doll{\'a}r}]{kirillov2019panoptic}
\bibinfo{author}{Kirillov, A.}, \bibinfo{author}{He, K.}, \bibinfo{author}{Girshick, R.}, \bibinfo{author}{Rother, C.}, \bibinfo{author}{Doll{\'a}r, P.}, \bibinfo{year}{2019}.
\newblock \bibinfo{title}{Panoptic segmentation}, in: \bibinfo{booktitle}{Proceedings of the IEEE/CVF conference on computer vision and pattern recognition}, pp. \bibinfo{pages}{9404--9413}.
\bibitem[{Kumar et~al.(2017)Kumar, Verma, Sharma, Bhargava, Vahadane and Sethi}]{kumar2017dataset}
\bibinfo{author}{Kumar, N.}, \bibinfo{author}{Verma, R.}, \bibinfo{author}{Sharma, S.}, \bibinfo{author}{Bhargava, S.}, \bibinfo{author}{Vahadane, A.}, \bibinfo{author}{Sethi, A.}, \bibinfo{year}{2017}.
\newblock \bibinfo{title}{A dataset and a technique for generalized nuclear segmentation for computational pathology}.
\newblock \bibinfo{journal}{IEEE transactions on medical imaging} \bibinfo{volume}{36}, \bibinfo{pages}{1550--1560}.
\bibitem[{Lai et~al.(2021)Lai, Tian, Jiang, Liu, Zhao, Wang and Jia}]{lai2021semi}
\bibinfo{author}{Lai, X.}, \bibinfo{author}{Tian, Z.}, \bibinfo{author}{Jiang, L.}, \bibinfo{author}{Liu, S.}, \bibinfo{author}{Zhao, H.}, \bibinfo{author}{Wang, L.}, \bibinfo{author}{Jia, J.}, \bibinfo{year}{2021}.
\newblock \bibinfo{title}{Semi-supervised semantic segmentation with directional context-aware consistency}, in: \bibinfo{booktitle}{Proceedings of the IEEE/CVF Conference on Computer Vision and Pattern Recognition}, pp. \bibinfo{pages}{1205--1214}.
\bibitem[{LaTorre et~al.(2013)LaTorre, Alonso-Nanclares, Muelas, Pe{\~n}a and DeFelipe}]{latorre2013segmentation}
\bibinfo{author}{LaTorre, A.}, \bibinfo{author}{Alonso-Nanclares, L.}, \bibinfo{author}{Muelas, S.}, \bibinfo{author}{Pe{\~n}a, J.}, \bibinfo{author}{DeFelipe, J.}, \bibinfo{year}{2013}.
\newblock \bibinfo{title}{Segmentation of neuronal nuclei based on clump splitting and a two-step binarization of images}.
\newblock \bibinfo{journal}{Expert Systems with Applications} \bibinfo{volume}{40}, \bibinfo{pages}{6521--6530}.
\bibitem[{Lin et~al.(2016)Lin, Dai, Jia, He and Sun}]{lin2016scribblesup}
\bibinfo{author}{Lin, D.}, \bibinfo{author}{Dai, J.}, \bibinfo{author}{Jia, J.}, \bibinfo{author}{He, K.}, \bibinfo{author}{Sun, J.}, \bibinfo{year}{2016}.
\newblock \bibinfo{title}{Scribblesup: Scribble-supervised convolutional networks for semantic segmentation}, in: \bibinfo{booktitle}{Proceedings of the IEEE conference on computer vision and pattern recognition}, pp. \bibinfo{pages}{3159--3167}.
\bibitem[{Liu et~al.(2021a)Liu, Zhang, Song, Huang and Cai}]{liu2021panoptic}
\bibinfo{author}{Liu, D.}, \bibinfo{author}{Zhang, D.}, \bibinfo{author}{Song, Y.}, \bibinfo{author}{Huang, H.}, \bibinfo{author}{Cai, W.}, \bibinfo{year}{2021}a.
\newblock \bibinfo{title}{Panoptic feature fusion net: a novel instance segmentation paradigm for biomedical and biological images}.
\newblock \bibinfo{journal}{IEEE Transactions on Image Processing} \bibinfo{volume}{30}, \bibinfo{pages}{2045--2059}.
\bibitem[{Liu et~al.(2021b)Liu, Guo, Cao and Tang}]{liu2021mdc}
\bibinfo{author}{Liu, X.}, \bibinfo{author}{Guo, Z.}, \bibinfo{author}{Cao, J.}, \bibinfo{author}{Tang, J.}, \bibinfo{year}{2021}b.
\newblock \bibinfo{title}{Mdc-net: A new convolutional neural network for nucleus segmentation in histopathology images with distance maps and contour information}.
\newblock \bibinfo{journal}{Computers in Biology and Medicine} \bibinfo{volume}{135}, \bibinfo{pages}{104543}.
\bibitem[{Loshchilov and Hutter(2017)}]{loshchilov2017decoupled}
\bibinfo{author}{Loshchilov, I.}, \bibinfo{author}{Hutter, F.}, \bibinfo{year}{2017}.
\newblock \bibinfo{title}{Decoupled weight decay regularization}.
\newblock \bibinfo{journal}{arXiv preprint arXiv:1711.05101} .
\bibitem[{M{\"u}ller et~al.(2019)M{\"u}ller, Kornblith and Hinton}]{muller2019does}
\bibinfo{author}{M{\"u}ller, R.}, \bibinfo{author}{Kornblith, S.}, \bibinfo{author}{Hinton, G.E.}, \bibinfo{year}{2019}.
\newblock \bibinfo{title}{When does label smoothing help?}
\newblock \bibinfo{journal}{Advances in neural information processing systems} \bibinfo{volume}{32}.
\bibitem[{Nam et~al.(2023)Nam, Jeong, Luna, Chikontwe and Park}]{nam2023pronet}
\bibinfo{author}{Nam, S.}, \bibinfo{author}{Jeong, J.}, \bibinfo{author}{Luna, M.}, \bibinfo{author}{Chikontwe, P.}, \bibinfo{author}{Park, S.H.}, \bibinfo{year}{2023}.
\newblock \bibinfo{title}{Pronet: Point refinement using shape-guided offset map for nuclei instance segmentation}, in: \bibinfo{booktitle}{International Conference on Medical Image Computing and Computer-Assisted Intervention}, \bibinfo{organization}{Springer}. pp. \bibinfo{pages}{528--538}.
\bibitem[{Naylor et~al.(2018)Naylor, La{\'e}, Reyal and Walter}]{naylor2018segmentation}
\bibinfo{author}{Naylor, P.}, \bibinfo{author}{La{\'e}, M.}, \bibinfo{author}{Reyal, F.}, \bibinfo{author}{Walter, T.}, \bibinfo{year}{2018}.
\newblock \bibinfo{title}{Segmentation of nuclei in histopathology images by deep regression of the distance map}.
\newblock \bibinfo{journal}{IEEE transactions on medical imaging} \bibinfo{volume}{38}, \bibinfo{pages}{448--459}.
\bibitem[{Niu et~al.(2024)Niu, Wang, Han, Lian, Herzig and Darrell}]{niu2024unsupervised}
\bibinfo{author}{Niu, D.}, \bibinfo{author}{Wang, X.}, \bibinfo{author}{Han, X.}, \bibinfo{author}{Lian, L.}, \bibinfo{author}{Herzig, R.}, \bibinfo{author}{Darrell, T.}, \bibinfo{year}{2024}.
\newblock \bibinfo{title}{Unsupervised universal image segmentation}, in: \bibinfo{booktitle}{Proceedings of the IEEE/CVF Conference on Computer Vision and Pattern Recognition}, pp. \bibinfo{pages}{22744--22754}.
\bibitem[{Oord et~al.(2018)Oord, Li and Vinyals}]{oord2018representation}
\bibinfo{author}{Oord, A.v.d.}, \bibinfo{author}{Li, Y.}, \bibinfo{author}{Vinyals, O.}, \bibinfo{year}{2018}.
\newblock \bibinfo{title}{Representation learning with contrastive predictive coding}.
\newblock \bibinfo{journal}{arXiv preprint arXiv:1807.03748} .
\bibitem[{Papandreou et~al.(2015)Papandreou, Chen, Murphy and Yuille}]{papandreou2015weakly}
\bibinfo{author}{Papandreou, G.}, \bibinfo{author}{Chen, L.C.}, \bibinfo{author}{Murphy, K.P.}, \bibinfo{author}{Yuille, A.L.}, \bibinfo{year}{2015}.
\newblock \bibinfo{title}{Weakly-and semi-supervised learning of a deep convolutional network for semantic image segmentation}, in: \bibinfo{booktitle}{Proceedings of the IEEE international conference on computer vision}, pp. \bibinfo{pages}{1742--1750}.
\bibitem[{Pathak et~al.(2015)Pathak, Krahenbuhl and Darrell}]{pathak2015constrained}
\bibinfo{author}{Pathak, D.}, \bibinfo{author}{Krahenbuhl, P.}, \bibinfo{author}{Darrell, T.}, \bibinfo{year}{2015}.
\newblock \bibinfo{title}{Constrained convolutional neural networks for weakly supervised segmentation}, in: \bibinfo{booktitle}{Proceedings of the IEEE international conference on computer vision}, pp. \bibinfo{pages}{1796--1804}.
\bibitem[{Qu et~al.(2020)Qu, Wu, Huang, Yi, Yan, Li, Riedlinger, De, Zhang and Metaxas}]{qu2020weakly}
\bibinfo{author}{Qu, H.}, \bibinfo{author}{Wu, P.}, \bibinfo{author}{Huang, Q.}, \bibinfo{author}{Yi, J.}, \bibinfo{author}{Yan, Z.}, \bibinfo{author}{Li, K.}, \bibinfo{author}{Riedlinger, G.M.}, \bibinfo{author}{De, S.}, \bibinfo{author}{Zhang, S.}, \bibinfo{author}{Metaxas, D.N.}, \bibinfo{year}{2020}.
\newblock \bibinfo{title}{Weakly supervised deep nuclei segmentation using partial points annotation in histopathology images}.
\newblock \bibinfo{journal}{IEEE transactions on medical imaging} \bibinfo{volume}{39}, \bibinfo{pages}{3655--3666}.
\bibitem[{Qu et~al.(2019)Qu, Yan, Riedlinger, De and Metaxas}]{qu2019improving}
\bibinfo{author}{Qu, H.}, \bibinfo{author}{Yan, Z.}, \bibinfo{author}{Riedlinger, G.M.}, \bibinfo{author}{De, S.}, \bibinfo{author}{Metaxas, D.N.}, \bibinfo{year}{2019}.
\newblock \bibinfo{title}{Improving nuclei/gland instance segmentation in histopathology images by full resolution neural network and spatial constrained loss}, in: \bibinfo{booktitle}{Medical Image Computing and Computer Assisted Intervention--MICCAI 2019: 22nd International Conference, Shenzhen, China, October 13--17, 2019, Proceedings, Part I 22}, \bibinfo{organization}{Springer}. pp. \bibinfo{pages}{378--386}.
\bibitem[{Rajchl et~al.(2016)Rajchl, Lee, Oktay, Kamnitsas, Passerat-Palmbach, Bai, Damodaram, Rutherford, Hajnal, Kainz et~al.}]{rajchl2016deepcut}
\bibinfo{author}{Rajchl, M.}, \bibinfo{author}{Lee, M.C.}, \bibinfo{author}{Oktay, O.}, \bibinfo{author}{Kamnitsas, K.}, \bibinfo{author}{Passerat-Palmbach, J.}, \bibinfo{author}{Bai, W.}, \bibinfo{author}{Damodaram, M.}, \bibinfo{author}{Rutherford, M.A.}, \bibinfo{author}{Hajnal, J.V.}, \bibinfo{author}{Kainz, B.}, et~al., \bibinfo{year}{2016}.
\newblock \bibinfo{title}{Deepcut: Object segmentation from bounding box annotations using convolutional neural networks}.
\newblock \bibinfo{journal}{IEEE transactions on medical imaging} \bibinfo{volume}{36}, \bibinfo{pages}{674--683}.
\bibitem[{Raza et~al.(2019)Raza, Cheung, Shaban, Graham, Epstein, Pelengaris, Khan and Rajpoot}]{raza2019micro}
\bibinfo{author}{Raza, S.E.A.}, \bibinfo{author}{Cheung, L.}, \bibinfo{author}{Shaban, M.}, \bibinfo{author}{Graham, S.}, \bibinfo{author}{Epstein, D.}, \bibinfo{author}{Pelengaris, S.}, \bibinfo{author}{Khan, M.}, \bibinfo{author}{Rajpoot, N.M.}, \bibinfo{year}{2019}.
\newblock \bibinfo{title}{Micro-net: A unified model for segmentation of various objects in microscopy images}.
\newblock \bibinfo{journal}{Medical image analysis} \bibinfo{volume}{52}, \bibinfo{pages}{160--173}.
\bibitem[{Ronneberger et~al.(2015)Ronneberger, Fischer and Brox}]{ronneberger2015u}
\bibinfo{author}{Ronneberger, O.}, \bibinfo{author}{Fischer, P.}, \bibinfo{author}{Brox, T.}, \bibinfo{year}{2015}.
\newblock \bibinfo{title}{U-net: Convolutional networks for biomedical image segmentation}, in: \bibinfo{booktitle}{Medical image computing and computer-assisted intervention--MICCAI 2015: 18th international conference, Munich, Germany, October 5-9, 2015, proceedings, part III 18}, \bibinfo{organization}{Springer}. pp. \bibinfo{pages}{234--241}.
\bibitem[{Sheeba et~al.(2014)Sheeba, Thamburaj, Mammen and Nagar}]{sheeba2014splitting}
\bibinfo{author}{Sheeba, F.}, \bibinfo{author}{Thamburaj, R.}, \bibinfo{author}{Mammen, J.J.}, \bibinfo{author}{Nagar, A.K.}, \bibinfo{year}{2014}.
\newblock \bibinfo{title}{Splitting of overlapping cells in peripheral blood smear images by concavity analysis}, in: \bibinfo{booktitle}{Combinatorial Image Analysis: 16th International Workshop, IWCIA 2014, Brno, Czech Republic, May 28-30, 2014. Proceedings 16}, \bibinfo{organization}{Springer}. pp. \bibinfo{pages}{238--249}.
\bibitem[{Sun et~al.(2024)Sun, Li, Torr, Gu and Li}]{sun2024clip}
\bibinfo{author}{Sun, S.}, \bibinfo{author}{Li, R.}, \bibinfo{author}{Torr, P.}, \bibinfo{author}{Gu, X.}, \bibinfo{author}{Li, S.}, \bibinfo{year}{2024}.
\newblock \bibinfo{title}{Clip as rnn: Segment countless visual concepts without training endeavor}, in: \bibinfo{booktitle}{Proceedings of the IEEE/CVF Conference on Computer Vision and Pattern Recognition}, pp. \bibinfo{pages}{13171--13182}.
\bibitem[{Upschulte et~al.(2022)Upschulte, Harmeling, Amunts and Dickscheid}]{upschulte2022contour}
\bibinfo{author}{Upschulte, E.}, \bibinfo{author}{Harmeling, S.}, \bibinfo{author}{Amunts, K.}, \bibinfo{author}{Dickscheid, T.}, \bibinfo{year}{2022}.
\newblock \bibinfo{title}{Contour proposal networks for biomedical instance segmentation}.
\newblock \bibinfo{journal}{Medical image analysis} \bibinfo{volume}{77}, \bibinfo{pages}{102371}.
\bibitem[{Vu et~al.(2019)Vu, Graham, Kurc, To, Shaban, Qaiser, Koohbanani, Khurram, Kalpathy-Cramer, Zhao et~al.}]{vu2019methods}
\bibinfo{author}{Vu, Q.D.}, \bibinfo{author}{Graham, S.}, \bibinfo{author}{Kurc, T.}, \bibinfo{author}{To, M.N.N.}, \bibinfo{author}{Shaban, M.}, \bibinfo{author}{Qaiser, T.}, \bibinfo{author}{Koohbanani, N.A.}, \bibinfo{author}{Khurram, S.A.}, \bibinfo{author}{Kalpathy-Cramer, J.}, \bibinfo{author}{Zhao, T.}, et~al., \bibinfo{year}{2019}.
\newblock \bibinfo{title}{Methods for segmentation and classification of digital microscopy tissue images}.
\newblock \bibinfo{journal}{Frontiers in bioengineering and biotechnology} \bibinfo{volume}{7}, \bibinfo{pages}{433738}.
\bibitem[{Wang et~al.(2021a)Wang, Zhou, Yu, Dai, Konukoglu and Van~Gool}]{wang2021exploring}
\bibinfo{author}{Wang, W.}, \bibinfo{author}{Zhou, T.}, \bibinfo{author}{Yu, F.}, \bibinfo{author}{Dai, J.}, \bibinfo{author}{Konukoglu, E.}, \bibinfo{author}{Van~Gool, L.}, \bibinfo{year}{2021}a.
\newblock \bibinfo{title}{Exploring cross-image pixel contrast for semantic segmentation}, in: \bibinfo{booktitle}{Proceedings of the IEEE/CVF international conference on computer vision}, pp. \bibinfo{pages}{7303--7313}.
\bibitem[{Wang et~al.(2021b)Wang, Zhang, Shen, Kong and Li}]{wang2021dense}
\bibinfo{author}{Wang, X.}, \bibinfo{author}{Zhang, R.}, \bibinfo{author}{Shen, C.}, \bibinfo{author}{Kong, T.}, \bibinfo{author}{Li, L.}, \bibinfo{year}{2021}b.
\newblock \bibinfo{title}{Dense contrastive learning for self-supervised visual pre-training}, in: \bibinfo{booktitle}{Proceedings of the IEEE/CVF conference on computer vision and pattern recognition}, pp. \bibinfo{pages}{3024--3033}.
\bibitem[{Xie et~al.(2021)Xie, Lin, Zhang, Cao, Lin and Hu}]{xie2021propagate}
\bibinfo{author}{Xie, Z.}, \bibinfo{author}{Lin, Y.}, \bibinfo{author}{Zhang, Z.}, \bibinfo{author}{Cao, Y.}, \bibinfo{author}{Lin, S.}, \bibinfo{author}{Hu, H.}, \bibinfo{year}{2021}.
\newblock \bibinfo{title}{Propagate yourself: Exploring pixel-level consistency for unsupervised visual representation learning}, in: \bibinfo{booktitle}{Proceedings of the IEEE/CVF conference on computer vision and pattern recognition}, pp. \bibinfo{pages}{16684--16693}.
\bibitem[{Yang et~al.(2018)Yang, Zhang, Zhao, Zheng, Liang, Ying, Ahuja and Chen}]{yang2018boxnet}
\bibinfo{author}{Yang, L.}, \bibinfo{author}{Zhang, Y.}, \bibinfo{author}{Zhao, Z.}, \bibinfo{author}{Zheng, H.}, \bibinfo{author}{Liang, P.}, \bibinfo{author}{Ying, M.T.}, \bibinfo{author}{Ahuja, A.T.}, \bibinfo{author}{Chen, D.Z.}, \bibinfo{year}{2018}.
\newblock \bibinfo{title}{Boxnet: Deep learning based biomedical image segmentation using boxes only annotation}.
\newblock \bibinfo{journal}{arXiv preprint arXiv:1806.00593} .
\bibitem[{Zhao et~al.(2018)Zhao, Yang, Zheng, Guldner, Zhang and Chen}]{zhao2018deep}
\bibinfo{author}{Zhao, Z.}, \bibinfo{author}{Yang, L.}, \bibinfo{author}{Zheng, H.}, \bibinfo{author}{Guldner, I.H.}, \bibinfo{author}{Zhang, S.}, \bibinfo{author}{Chen, D.Z.}, \bibinfo{year}{2018}.
\newblock \bibinfo{title}{Deep learning based instance segmentation in 3d biomedical images using weak annotation}, in: \bibinfo{booktitle}{Medical Image Computing and Computer Assisted Intervention--MICCAI 2018: 21st International Conference, Granada, Spain, September 16-20, 2018, Proceedings, Part IV 11}, \bibinfo{organization}{Springer}. pp. \bibinfo{pages}{352--360}.
\bibitem[{Zhou et~al.(2022)Zhou, Nie, Adeli, Wei, Ren, Liu, Zhu, Yin, Wang and Shen}]{zhou2022semantic}
\bibinfo{author}{Zhou, S.}, \bibinfo{author}{Nie, D.}, \bibinfo{author}{Adeli, E.}, \bibinfo{author}{Wei, Q.}, \bibinfo{author}{Ren, X.}, \bibinfo{author}{Liu, X.}, \bibinfo{author}{Zhu, E.}, \bibinfo{author}{Yin, J.}, \bibinfo{author}{Wang, Q.}, \bibinfo{author}{Shen, D.}, \bibinfo{year}{2022}.
\newblock \bibinfo{title}{Semantic instance segmentation with discriminative deep supervision for medical images}.
\newblock \bibinfo{journal}{Medical Image Analysis} \bibinfo{volume}{82}, \bibinfo{pages}{102626}.
\bibitem[{Zhou et~al.(2019)Zhou, Onder, Dou, Tsougenis, Chen and Heng}]{zhou2019cia}
\bibinfo{author}{Zhou, Y.}, \bibinfo{author}{Onder, O.F.}, \bibinfo{author}{Dou, Q.}, \bibinfo{author}{Tsougenis, E.}, \bibinfo{author}{Chen, H.}, \bibinfo{author}{Heng, P.A.}, \bibinfo{year}{2019}.
\newblock \bibinfo{title}{Cia-net: Robust nuclei instance segmentation with contour-aware information aggregation}, in: \bibinfo{booktitle}{Information Processing in Medical Imaging: 26th International Conference, IPMI 2019, Hong Kong, China, June 2--7, 2019, Proceedings 26}, \bibinfo{organization}{Springer}. pp. \bibinfo{pages}{682--693}.

\end{thebibliography}



\end{document}